\documentclass{article}

\usepackage{microtype}
\usepackage{graphicx}
\usepackage{subfigure}
\usepackage{booktabs} 

\usepackage{amsfonts}       
\usepackage{amsmath}
\usepackage{nicefrac}       
\usepackage{microtype}      
\usepackage{graphicx}
\usepackage{wrapfig}
\usepackage[ruled,vlined,algo2e]{algorithm2e}
\usepackage{enumitem}
\usepackage{amsthm}
\usepackage{todonotes}
\usepackage{comment}
\usepackage{multicol}
\usepackage{wrapfig}
\usepackage{mathtools}

\usepackage{hyperref}

\newtheorem{thm}{Theorem}
\newtheorem{remark}[thm]{Remark}

\usepackage[accepted]{util/icml2020}

\icmltitlerunning{\color{white}No Representation without Transformation}

\begin{document}

\twocolumn[
\icmltitle{No Representation without Transformation}

\icmlkeywords{Latent Variable Models, Transformations}
\begin{icmlauthorlist}
\icmlauthor{Giorgio Giannone}{nn}
\icmlauthor{Saeed Saremi}{nn,uc}
\icmlauthor{Jonathan Masci}{nn}
\icmlauthor{Christian Osendorfer}{nn}
\end{icmlauthorlist}

\icmlaffiliation{nn}{NNAISENSE, Lugano, Switzerland }
\icmlaffiliation{uc}{Redwood Center for Theoretical Neuroscience, UC Berkeley}
\icmlcorrespondingauthor{}{giorgio.c.giannone@gmail.com}
\icmlcorrespondingauthor{}{saeed@berkeley.edu}
\icmlcorrespondingauthor{}{christian@nnaisense.com}
\vskip 0.3 in
]



\printAffiliationsAndNotice{}  

\begin{abstract}
    We extend the framework of variational autoencoders to represent transformations explicitly in the latent space. 
    In the family of hierarchical graphical models that emerges, the latent space is populated by higher order objects that are inferred jointly with the latent representations they act on. 
    To explicitly demonstrate the effect of these higher order objects, we show that the inferred latent transformations reflect interpretable properties in the observation space. 
    Furthermore, the model is structured in such a way that in the absence of transformations, we can run inference and obtain generative capabilities comparable with standard variational autoencoders. 
    Finally, utilizing the trained encoder, we outperform the baselines by a wide margin on a challenging out-of-distribution classification task. 
\end{abstract}

\section{Introduction}
\label{sec:intro}

How we represent the world is intricately tied to how the world transforms. This idea has deep roots in mathematics~\citep{lie1871over, klein}, theoretical physics, at the birth of quantum mechanics~\citep{weyl1927quantenmechanik, wigner1931gruppentheorie}, Gestalt psychology, and also in artificial intelligence pioneered by David Marr's thesis on  \emph{object-centered coordinate system} for the problem of vision~\citep{marr1978representation, marr}. The first \emph{connectionist} realization of Marr's ideas was proposed by~\citet{hinton1981parallel} which outlined a (neural) network model for transforming reference frames with a set of ``mapping units'' to gate the connections between input and output units depending on  the transformation; 
that itself inspired research on neurobiological models under \emph{dynamic routing circuits}~\citep{anderson1987shifter, olshausen1995multiscale}. These old ideas on \emph{explicitly} representing transformations with \emph{hidden neurons} have had a resurgence recently
spurred by the renewed interest in neural network based approaches \citep{memisevic2007unsupervised, hinton2011transforming,memisevic2012multi,michalski2014modeling,cohen2014transformation, sabour2017dynamic}.

The language to describe transformations (and invariances under transformations) is \emph{group theory} and \emph{representation theory}~\citep{kowalski2014introduction} with its deep and long trace in the history of physics. The integration of group theoretical methods into machine learning is quite recent however and explored from different angles, in the thesis~\citep{kondor2008group} and~\citep{mallat2012group, bruna2013invariant}. Group theory methods have also made their way into constructing novel \emph{architectures} where representation theory is used to generalize convolutional neural networks~\citep{cohen2016group}. These are important developments in bringing representation theory into machine learning and signal processing.

In machine learning, we must deal with uncertainties in data. But in contrast with ``beautiful'' and symmetric physical theories, the data is in general quite ``messy'' which makes the use of representation theory  limited in principle with approximations that we must control. With that in mind, we formulate a notion of representation theory for probabilistic models where transformations are represented explicitly in the latent space of a latent variable model, together with an algebra on how they ``act'' on other latent variables: \begin{quote}{\it Our main conceptual contribution is this representation of group actions in the latent space that we integrate with variational autoencoders with the ultimate goal of unifying representation theory and representation learning. The family of models that emerge is referred to as transformation-aware variational autoencoders, $\mathcal{T}_{\rm VAE}$~for short. At the technical level, (i) a graphical model is set up
to represent transformations, (ii) we propose an approximation to the posterior over the latent variables and derive a variational lower bound,  (iii) guided by the expression for the lower bound, we relax some of the dependencies between latent variables to be deterministic, (iv) a learning objective emerges and we demonstrate 
empirically that the algebra of transformations can indeed be ``enforced'' in the
latent space in a probabilistic manner.}    \end{quote}

In $\mathcal{T}_{\rm VAE}$, the (complex) transformations $g \in G$ that act on the random variable $X$ in the observation space $\mathbb{R}^{d_X}$ are instead explicitly represented as higher order latent variables, denoted by $T$, which we can infer in the latent space together with the latent variables $Z$ that $T$ ``acts on''.  
The idea is that transformations are better behaved in the latent space\textemdash the action of $\tau \sim T$ on $z\sim Z$ is ``simpler'' than the action of $g$ on $x \sim X$\textemdash and perhaps, in the spirit of representation theory, they can even be approximated by linear transformations; in that case, in the jargon of representation theory, the vector space $\mathbb{R}^{d_Z}$ is (approximately) a representation of the group of transformations $G$.

In practice, the group $G$ and how it acts on the random variable $X$ in $\mathbb{R}^{d_X}$  can be unknown to us, except that we assume that elements in $G$ satisfy the general group properties under the binary operation on $G$~\citep{robinson2012course}. As implied in the statements above, the best we can do is to \emph{approximate} a representation of this group in the latent space, and that is our starting point. In $\mathcal{T}_{\rm VAE}$, the transformations $g\in G$ are not represented in the observation space $X$ but they become ``first class citizens'' in the form of probabilistic variables $T$ together with  an algebra on how they act on the latent variables $Z$ in $\mathbb{R}^{d_Z}$. All the random variables $(\{X\}, \{Z\}, \{T\})$ are put together in a graphical model that dictates their joint density:
\begin{itemize}
\item $\{X\}$ denotes the \emph{ordered set} that represents random variables $X$ and its transformations; $\{Z\}$ is the corresponding ordered set for the random variables $Z$. 
\item $T$ is viewed both as a random variable and as an operator, and the same goes for samples $\tau \sim T$. For example, $T$ can be a random $d_Z \times d_Z$ matrix whose samples $\tau \sim T$ acts on $z\sim Z$ as: $z\mapsto \tau z$. Or $\tau \sim T$ can be a vector in $\mathbb{R}^{d_Z}$ that acts on $z$ as $z\mapsto z + \tau$.
\end{itemize}

The framework of choice for learning the graphical model associated with $(\{X\}, \{Z\}, \{T\})$ is \emph{variational inference}~\citep{jordan1999introduction, wainwright2008graphical, hoffman2013stochastic} where $\log p(\{X\})$ is approximated with a \emph{variational lower bound}  and training the generative network (the decoder) and the inference network (the encoder) is achieved with \emph{amortized inference}~\citep{gershman2014amortized} and the ``reparametrization trick'' in \emph{variational autoencoders}~\citep{kingma2013auto, rezende2014stochastic}. Next, we present a graphical model for $(\{X\}, \{Z\}, \{T\})$ for which we derive a variational lower bound and a learning objective, and discuss the different instantiations of $\mathcal{T}_{\rm VAE}$ depending on the algebraic form of how $T$ acts on $Z$. 

\section{Transformation-aware VAE}
\label{sec:approach}

\begin{figure}[t!]
    \begin{center}
	\subfigure[$\mathrm{VAE}$]{
		\includegraphics[scale=0.7]{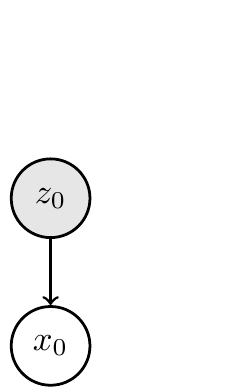}
	}
	\subfigure[$\mathrm{VAE}_{\rm id}$]{
		\includegraphics[scale=0.7]{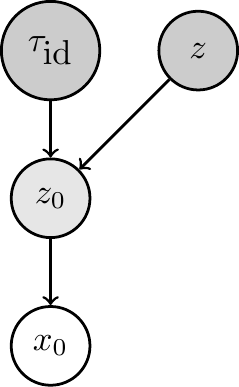}
	}
	\subfigure[$\mathcal{T}_{\mathrm{VAE}}$]{
		\includegraphics[scale=0.7]{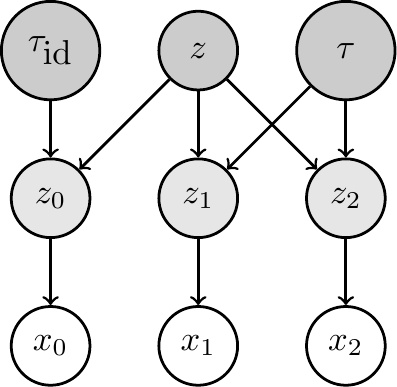}
	}
	\caption{
		(a) A standard VAE.
		(b) If no real latent transformation exists, the suggested model collapses to a standard VAE.
		(c) Our model, $\mathcal{T}_{\rm VAE}$.  
		A latent sample $z$ is transformed into two views, $z_1$ and $z_2$. 
		The identity transformation maps $z$ to $z_0$. These latent representations are then mapped through a directed graphical model (a decoder) to the observation space.}
	\label{fig:gm}
	\end{center}
    \vskip -0.2in
\end{figure}

What does it mean to endow the latent space of a variational autoencoder with a random variable $T$ representing latent transformations that act on $Z$? The framework of VAEs is used for describing directed probabilistic models for i.i.d. samples (see Figure~\ref{fig:gm}a for the graphical model). Of course we can postulate that a special transformation already exists in this latent space, the identity function $\tau_{\mathrm{id}}: z \mapsto z$ (see Figure~\ref{fig:gm}b). This construction seems cumbersome and unnecessary but  it gives an important insight: Integrating transformations in the latent space means expressing relationships between latent variables and leads to hierarchical directed models. 


How should this hierarchical structure look like? Samples from $T$ represent individual transformations that act in the latent space, so can we find a structure that represents properties the group $G$ has? A key property of the group $G$ is that each element $g$ has an inverse $g^{-1}$, $e = g \circ g^{-1}$. Therefore, one property of $T$ (being a group representation of $G$ under function composition) is that for every $\tau$ an \emph{inverse} $\tau^{-1}$ exists. One way to express this is to extend Figure~\ref{fig:gm}b by introducing two additional latent variables $z_1$ (samples from the random variable $Z_1$) and $z_2$ such that $z_1 = \tau(z)$ and $z_2 = \tau^{-1}(z)$ for $\tau \sim T$. 
\begin{remark}
	The expression $\tau(z)$ is a compact notation for the group action in the latent space as parameterized by $\tau$.
	\label{remark:tau_operator}
\end{remark}
While now the \emph{algebraic assumption of the existence of inverses} is encoded in the latent space, the latent variables $z_i$ still need to be grounded in the observation space, which is achieved by a standard directed probabilistic model $$p(x_i \vert z_i).$$ Taken as a whole the graphical model --- the \emph{transformation-aware} VAE $\mathcal{T}_{\rm VAE}$ --- that emerges is shown in Figure~\ref{fig:gm}c. Note that this model describes distributions over \emph{i.i.d.} \emph{triplets} $(x_0,x_1,x_2)$, but the samples within the triplets are clearly not statistically independent. These triplets reflect the construction in the latent space: $x_1$ and $x_2$ are transformed and inversely transformed views of $x_0$, that is transformations in the observation space are described \emph{implicitly}. Many interesting transformations in the observation space do not have analytical closed form representations, so this approach provides a lot of flexibility.

From Figure~\ref{fig:gm}c we can see that the conditional likelihood  
$p_{\theta, \chi}((x_0, x_1, x_2), (z_0, z_1, z_2) \vert z, \tau, \tau_{\mathrm{id}})$ 
factorizes as
\begin{equation}
\prod_{i=0}^2
p_{\theta}(x_i \vert z_i)
r_{\chi}(z_0 \vert z,\tau_{\mathrm{id}})
r_{\chi}(z_1 \vert z, \tau)
r_{\chi}(z_2 \vert z, \tau^{-1}).
\label{eq:condlog}
\end{equation}
The operational form of $r_\chi(z_1 \vert z, \tau)$ is not specified here; its role is to ``encode'' how $z_1$ is related to $\tau(z)$ (see Remark~\ref{remark:tau_operator}), the same goes for $r_{\chi}(z_0 \vert z,\tau_{\mathrm{id}})$ and $r_{\chi}(z_2 \vert z, \tau^{-1})$. In Section~\ref{sec:experiments} we explore three functionally different realizations of $r_\chi$. The full joint distribution is then given by
\begin{equation}
    p_{\theta, \chi}((x_0, x_1, x_2), (z_0, z_1, z_2) \vert z, \tau, \tau_{\mathrm{id}}) \pi(z)\pi(\tau)\delta(\tau_{\mathrm{id}}),
\end{equation}
where the prior $\delta(\tau_{\mathrm{id}})$ on the random variable $\tau_{\rm id}$ is considered to be a delta function centered at the identity of the transformation group.



Maximizing the log-likelihood $\log p_{\theta, \chi}((x_0, x_1, x_2))$ drives the learning but assumptions have to be made on the form of the posterior $p_{\theta, \chi}((z_0, z_1, z_2), z, \tau \vert (x_0, x_1, x_2))$. In this work, we consider the following approximate factorization:
\begin{equation}
q_{\xi}(\tau \vert z_1, z_2) 
q_{\psi}(z \vert z_1, z_2)
\prod_{i=0}^2
q_{\phi}(z_i \vert x_i).
\label{eq:post}
\end{equation}
What parametric distribution should $q_{\xi}(\tau \vert z_1, z_2)$ be? Due to the hierarchical structure of the graphical model, and to have control over the allocation of uncertainty, we opted for a deterministic mapping parametrized by $\xi$: \begin{equation} \label{eq:tau} \tau = f_\xi(z_1,z_2).\end{equation}
With this setup, we arrive at the following expression for the learning objective $\mathcal{L}(\theta, \phi, \psi, \chi, \xi)$:
\begin{align}
	\begin{split}
   \hspace{-8pt} &\underset{{q_{\phi}(z_0 \vert x_0)}}{\mathbb{E}}\big( 
	\log p_{\theta}(x_0 \vert z_0)
    -\mathcal{D}\left[q_{\phi}(z_0 \vert x_0), r_{\chi}(z_0 \vert z, \tau_{\mathrm{id}}) \right]\big)\\
  +\hspace{-8pt} &\underset{q_{\phi}(z_1 \vert x_1)}{\mathbb{E}}
    \log p_{\theta}(x_1 \vert z_1)
	- \mathcal{D}\left[q_{\phi}(z_1 \vert x_1), r_{\chi}(z_1 \vert z, \tau) \right] \\
  +\hspace{-8pt} &\underset{q_{\phi}(z_2 \vert x_2)}{\mathbb{E}}
	\log p_{\theta}(x_2 \vert z_2)
	- \mathcal{D}\left[q_{\phi}(z_2 \vert x_2), r_{\chi}(z_2 \vert z, \tau^{-1}) \right] \\
 -\hspace{-8pt} & \quad \mathcal{D}
	 \left[ q_{\psi}(z \vert z_1, z_2),  \pi(z)
	 \right],
	\end{split}
	\label{eq:lb}
\end{align}
where
\begin{equation}
\mathcal{D}\left[a, b\right]=
\mathbb{E}_{
    q_{\psi}(z \vert z_1, z_2)
    q_{\phi}(z_1 \vert x_1) 
    q_{\phi}(z_2 \vert x_2)
    } \log a / b .
\label{eq:div}
\end{equation}

In Equation~\ref{eq:lb}, $\tau$ is \emph{computed} (deterministically) as stated in Eq.~\ref{eq:tau}, where $z_1$ and $z_2$ are themselves sampled from $q(z_1 \vert x_1)$ and $q(z_2 \vert x_2)$
respectively. 
See Appendix A for a full derivation of the lower bound that the expression for the learning objective $\mathcal{L}(\theta, \phi, \psi, \chi, \xi)$ in Eq.~\ref{eq:lb} is based on.

\begin{remark}
Inspecting the objective, we see that $$ \mathcal{D}\left[q_{\phi}(z_1 \vert x_1), r_{\chi}(z_1 \vert z, \tau) \right] $$
enforces the desired property that samples from the posterior of the transformed data $q_{\phi}(z_1 \vert x_1)$ should match the transformations of the latent variable $z$ as dictated by $r_{\chi}(z_1 \vert z, \tau)$. It therefore bridges the group actions in the observation space to the corresponding transformations in the latent space.
\label{remark:div}
\end{remark}

\begin{remark}
Eq.~\ref{eq:condlog} and Eq.~\ref{eq:post} are intentionally structured in the way presented. This construction allows to \emph{extract} a VAE-like encoder-decoder architecture after training a $\mathcal{T}_{\rm VAE}$. 
\end{remark}

\section{Results}
\label{sec:experiments}
In this section we describe the augmentation used on the datasets to generate triplets, implementation details of the presented mathematical formulation for $\mathcal{T}_{\textrm{VAE}}$, and three sets of experiments to validate our approach.

\begin{figure}[t!]
	\vskip 0.2in
    \begin{center}
	\subfigure[MNIST]{
    \includegraphics[width=.3\linewidth]{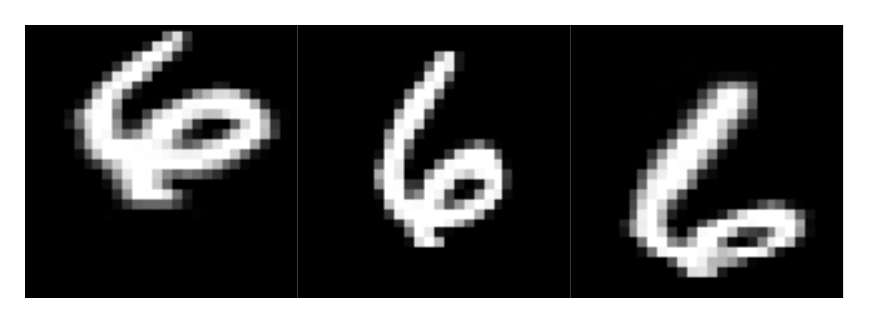}
    }
    \subfigure[F-MNIST]{
    \includegraphics[width=.3\linewidth]{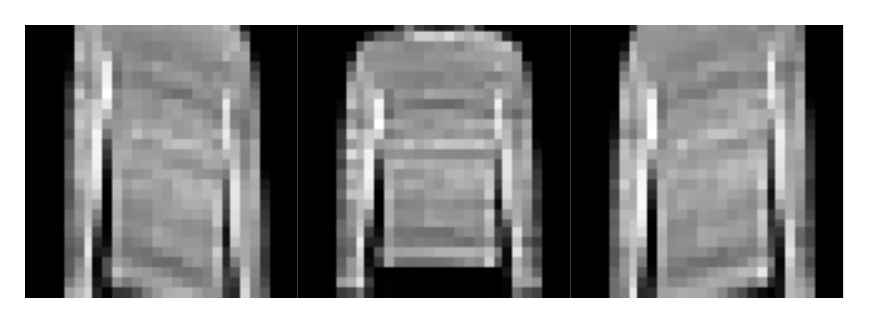}
    }
    \subfigure[OMNIGLOT]{
    \includegraphics[width=.3\linewidth]{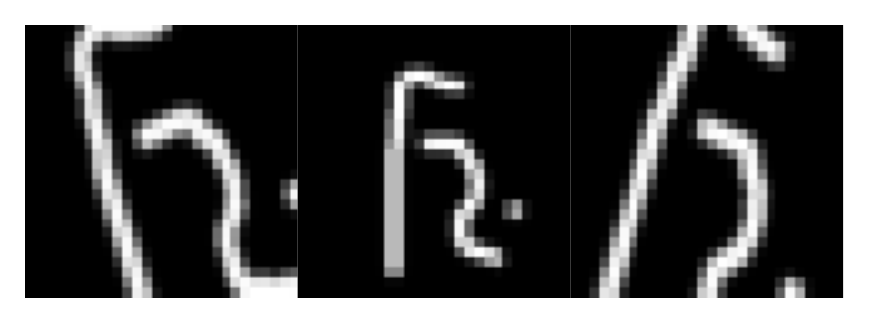}
    }
    \caption{Sampled triplets with randomly applied transformations. 
    The central element is from the original dataset; the left and right images are generated by random transformations. 
    These two transformations are approximately inverse to each other.} 
    \label{fig:samples}
    \end{center}
    \vskip -0.2in
\end{figure}

\subsection{Dataset augmentation}
We generate triplets by applying three different types of (class-preserving) transformations: size preserving rotation (rotation-cropping-zoom), tilting and shear.\footnote{We modify the \href{https://github.com/mdbloice/Augmentor}{Augmentor} library to augment the data.} 
When sampling triplets for training, these transformations are applied on the fly to the original data samples. A triplet needs two transformations given a sample from the base dataset. We choose the two transformations so that they are (almost) inverse to each other (due to the nonlinear cropping transformation exact inverses are not possible): The first transformation is chosen randomly, and the second transformation, i.e. the inverse, is determined from the first one. In general we choose transformations that satisfy two criteria: (a) they distort the sample substantially (translations by few pixels are not ``substantial'') --- this means that we generate a distributional shift in the data domain. Here, a dataset is considered \textit{domain shifted} by a set of transformations $G$ if a model trained on the same data without the transformations cannot recognize or reconstruct the transformed samples. (b) they are geometrically and physically plausible, e.g. an MNIST digit is not scaled beyond the size of the original image. See Figure \ref{fig:samples} for triplet examples generated in the described way.

\subsection{Implementation}
\label{subsec:model_details}
While the algebraic properties of the latent mapping allowed us to derive the model above, several more details are required to instantiate the model presented in Section~\ref{sec:approach} and Figure~\ref{fig:gm}c. Specifically we explore different $r_{\chi}$ parameterizations and also present an implementation of the divergence in Eq.~\ref{eq:div}.

There are many reasonable ways to represent the action of a transformation on a latent representation.
To assess the overall idea, we decided to use three different functional forms for $r_{\chi}$ that are representative for a large class of possible transformations and different levels of expressivity:

\begin{itemize}
    \item[$-$] \emph{Additive} (A)\footnote{A, M, N are used as abbreviations in tables.}: $\tau$ is parameterized by a vector in $\mathbb{R}^{d_Z}$. For $\tau(z)$
    we use the simple functional form of a translation for the latent transformation 
    $r_{\chi}(z_1 \vert z, \tau) = \delta(z_1-(z + \tau))$, i.e. $z_1$ is set to be equal to $z+\tau$ in the graphical model. 
    Its inverse $r_{\chi}(z_2 \vert z, \tau^{-1}) = \delta(z_2 -(z - \tau))$, i.e. $z_2$ is set to be equal to $z-\tau$. 
    \item[$-$] \emph{Matrix} (M): $\tau$ is parameterized by either a block diagonal matrix or by a tri-diagonal matrix. 
    In the block diagonal case, each block is represented as 2d rotational matrices $M(\theta)$. 
    Therefore, for $k = \tfrac{d_Z}{2}$,
        $$\tau(\theta_1,\dots,
    \theta_k) = \begin{psmallmatrix}
        M(\theta_1) &  & 0 \\
         & \ddots  & \\
         0 & & M(\theta_k)  \end{psmallmatrix},
        $$
    where $d_Z$ is chosen to be even, therefore the matrix $\tau$ has the dimensions $d_Z \times d_Z$. Here, $r_{\chi}(z_1 \vert z, \tau) = \delta(z_1  - \tau z)$ and
    $r_{\chi}(z_2 \vert z, \tau^{-1}) =\delta(z_2 - \tau^{-1} z))$, where $\tau^{-1}(\theta_1,\theta_2, ...) = \tau(-\theta_1, -\theta_2, ...)$.
    
    Inspired by the form of the above matrix we also experimented with general tri-diagonal matrices, where $\tau^{-1}$ was approximated by its transpose. Results for this parameterization are only shown in Appendix D, and are denoted by T there.
    \item[$-$] \emph{Neural} (N): The most flexible representation for $\tau(z)$ is by a Neural Network, $\textrm{NN}_{\tau}(z)$. $\textrm{NN}_{\tau}(z)$ hereby describes a neural network that is parameterized by the random variable $\tau$. In our experiments we choose an input parameterization, i.e. $\textrm{NN}_{\tau}(z) \equiv \textrm{NN}(z, \tau)$, a neural network with two input values. The dimension of the latent variable $\tau$ in this case can be arbitrary, we tied it to the dimension of $z$. That is,
    $$r_{\chi}(z_1 \vert z, \tau) = \delta(z_1 - (z + \text{NN}(z, \tau))),$$ 
    $$r_{\chi}(z_2 \vert z, \tau^{-1}) =\delta(z_2 - (z + \text{NN}(z, -\tau))),$$
\end{itemize}

The delta functions as the functional form for $r_\chi$ are motivated by ``forcing'' a specific algebraic form on $z_1$ and $z_2$, but a practical problem with delta functions is computing the divergence terms $\mathcal{D}[a, b]$ (Eq.~\ref{eq:div}) due to the presence of the logarithms which ``blow up'' when the ``constraints'' enforced by the delta function are not satisfied. To overcome this issue we substitute the divergence terms involving $r_\chi$ with Maximum Mean Discrepancy (MMD)  \citep{gretton2007kernel} using a linear kernel\textemdash also see~\citep{ tolstikhin2017wasserstein, zhao2017infovae} for the use of MMD for variational autoencoders. As an example, we replace $\mathcal{D}[q_\phi(z_1 \vert x_1), r_\chi(z_1 \vert z, \tau)]$ with 
\begin{equation}
 \left \Vert  \mathbb{E}_{q_\phi(z_1 \vert x_1)}[z_1] - \mathbb{E}_{r_\chi(z_1 \vert z, \tau)}[ z_1] \right \Vert^2,
 \label{eq:mmd}
\end{equation}
where, for the right term,  $\tau$ is computed by $q_\xi(\tau \vert z_1, z_2)$ and $z$ is sampled from $q_\psi(z \vert z_1, z_2)q_\phi(z_1 \vert x_1)q_\phi(z_2 \vert x_2)$. 

In our model, $z$ and $z_0$ should behave the same: they should give rise to the same likelihoods for $x$\textemdash this semantic is \emph{implicitly} enforced by the term 
\begin{equation}
\label{eq:D1} 
\mathcal{D}\left[q_{\phi}(z_0 \vert x_0), r_{\chi}(z_0 \vert z, \tau_{\mathrm{id}}) \right]
\end{equation} 
in Eq.~\ref{eq:lb} (see Remark~\ref{remark:div}). However, in practice, the reconstruction from the posterior $q_\psi(z|z_1, z_2)$ did not match the quality of the reconstruction from $q_\phi(z_0|x_0)$. As a solution to this issue we substitute $\mathcal{D}$, the expression in~(\ref{eq:D1}), with
\begin{equation}
\underset{
\substack{
    q_{\psi}(z \vert z_1, z_2)\\
    q_{\phi}(z_1 \vert x_1) 
    q_{\phi}(z_2 \vert x_2)
    }
    }
{\mathbb{E}} \hspace{-20pt}
- \log p_{\theta, \chi}(x \vert z)
+
\mathcal{D}_{\textrm{KL}} \left[q_{\phi}(z_0 \vert x_0), \pi(z_0)\right],
\label{eq:constraint}
\end{equation}
where the first term ensures that $x$ has a high likelihood under the posterior $q_\psi(z \vert z_1, z_2)$, and the second term, a Kullback-Leibler divergence, provides a proper prior for $z_0$. The resulting loss we actually use for training is presented in Appendix A.

Finally, for the posterior $q_{\psi}(z \vert z_1, z_2)$ we choose a Gaussian whose mean and standard deviation are parameterized by a neural network with parameters $\psi$. The implicit model $q_{\xi}(\tau \vert z_1, z_2)$ is parameterized by a neural network as well. The prior $\pi(\cdot)$ is the Gaussian $N(0, 1)$. We provide architectural details for all modules in Appendix E.

\subsection{Experiments}
With the following experiments we demonstrate that:
\begin{enumerate}[label=(\roman*)]
    \item Our model represents transformations in latent space, and 
          the random variable $T$ encodes information that resembles transformations in observation space.
    \item The resulting $\mathcal{T}_{\textrm{VAE}}$ model is also an effective generative model for the i.i.d dataset it was bootstrapped off.
    \item The latent structure from the $\mathcal{T}_{\textrm{VAE}}$ is qualitatively different from a normal VAE and is well-suited to handle out-of-sample data.
\end{enumerate}

In all experiments we use a standard VAE as a baseline. Because $\mathcal{T}_{\textrm{VAE}}$ is trained on triplets and hence on an artificially augmented dataset, we also introduce VAE+ as another baseline: a VAE trained on \emph{single samples} from an augmented dataset constructed with the same transformations as employed during training of a $\mathcal{T}_{\textrm{VAE}}$. These three models share the same architecture for the core encoder $q(z \vert x)$ and the core decoder $p(x \vert z)$ in all experiments. We run experiments on MNIST \cite{lecun1998mnist}, Fashion-MNIST \cite{xiao2017fashion}, Omniglot \cite{lake2015human} and AffNIST \cite{affNIST}.

\paragraph{Latent transformations.}
The main difference between $\mathcal{T}_{\textrm{VAE}}$ and other models based on the variational autoencoder framework is the latent variable $\tau$ representing transformations in the latent space. In the first set of experiments we want to investigate what information is encoded in $\tau$.

For these experiments we train the three variants of $\mathcal{T}_{\textrm{VAE}}$ on triplets generated from MNIST. After training is finished, we take an arbitrary triplet $(x_0, x_1, x_2)$ constructed according to our data generation scheme and infer $\tau$. We then take a new sample $x$ randomly from the test set, sample from its posterior $q_{\phi}(z \vert x)$, deterministically compute $z_1$ via $r_{\chi}(z_1 \vert z, \tau)$ and $z_2$ via $r_{\chi}(z_2 \vert z, \tau^{-1})$ and decode a new $x_1$ from $z_1$ and a new $x_2$ from $z_2$.

\begin{figure}[t!]
\vskip 0.2in
\begin{center}
\subfigure[]{
\centerline{\includegraphics[width=\columnwidth]{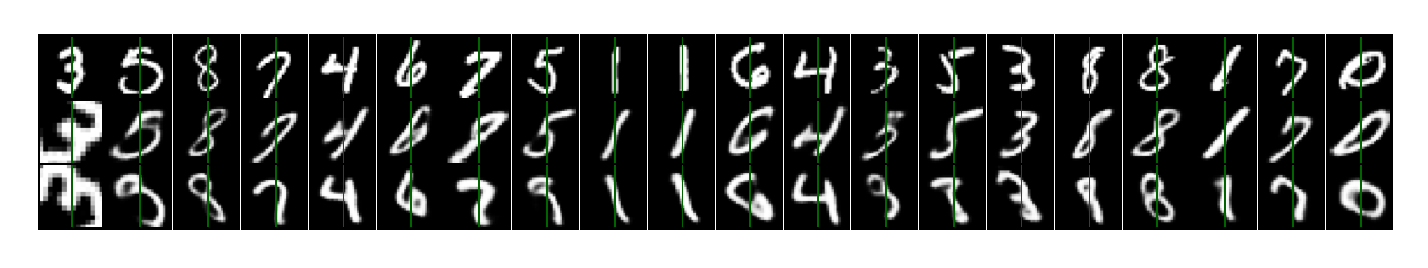}}}

\subfigure[]{
\centerline{\includegraphics[width=.9\columnwidth]{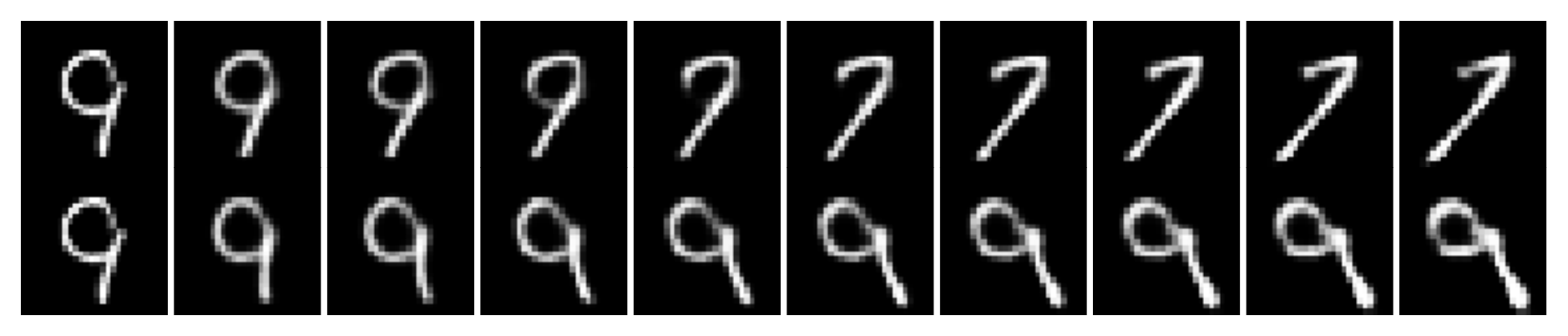}}}
\caption{(a) The transformation $\tau$, here a rotation, is extracted from the triplet in the first column. It is then applied to the latent embedding of the top image in the subsequent columns. The green vertical lines in every image should support inspecting for right/left rotation. (b) Repeatedly applying an inferred transformation. The first column shows the same number ("9") twice. In the next column, we apply a small rotation to the right (top), and it's inverse to the left (bottom row). $\tau$ is inferred from these two images (and the image in the first column), and then repeatedly applied onto the latent representations, generating the images in the subsequent columns (top image is generated by $\tau$, bottom image is generated by $\tau^{-1}$. Best viewed electronically.}
\label{fig:commutativity_paper}
\end{center}
\vskip -0.2in
\end{figure}

Figure~\ref{fig:commutativity_paper}a shows the qualitative result of this experiment for the \emph{Additive} parameterization of $r_\chi$. The triplet for extracting $\tau$ is in the first column: The top image shows an unmodified digit $3$ and the two images below show a right (i.e. $x_1$) and left (i.e. $x_2$) rotation of it respectively. For both of these images we can also see the effects of the (necessary) cropping transformation. This means that the two overall transformations are not true inverses to each other. Nevertheless, one would expect that the inferred $\tau$ should encode the rotation transformation. The following columns are then constructed by applying this inferred $\tau$: The images in the top row are randomly sampled test images, the second and the third row show the result of applying $\tau$ (second row) and $\tau^{-1}$(third row) to the latent embedding of the first row. We see that most examples are rotated to the right and left respectively while content and style are mostly unmodified. We also investigated the matrix and neural network variant of $r_\chi$ --- overall the results where qualitatively not as good as those shown in Figure~\ref{fig:commutativity_paper}a. 
These results are shown in Appendix D where we also explain one potential reason for these observations.

We also looked into a different experiment for the \emph{Additive} variant: What is happening when we repeatedly apply an inferred $\tau$? 
In Figure~\ref{fig:commutativity_paper}b this is shown for a rotation. The same trained $\mathcal{T}_{\textrm{VAE}}$ as previously is used. A small rotation to some image is applied and the underlying $\tau$ is inferred. The rest of the images in this figure are then generated by repeatedly applying this $\tau$ in latent space and decoding the respective latent representations into the observation space.

\paragraph{Generative modeling.} 
In the $\mathcal{T}_{\mathrm{VAE}}$, the encoder $q_\phi(z|x)$ and the decoder $p_\theta(x|z)$ can be combined to form a standard VAE. In this section we briefly investigate if a VAE constructed in this way has a reasonable generative capacity to model i.i.d samples.

We train the three variants of the $\mathcal{T}_{\mathrm{VAE}}$ on three different datasets (MNIST, F-MNIST and Omniglot) and report marginal log-likelihoods for the respective test set of these datasets. The marginal log-likelihoods are computed by importance sampling, using the trained encoder and decoder from the $\mathcal{T}_{\mathrm{VAE}}$ like in a standard VAE. We train a standard VAE and a data-augmented VAE+ as baselines, using the same architecture (details in Appendix E) for the encoders and decoders as for the $\mathcal{T}_{\mathrm{VAE}}$. Table~\ref{tab:generative-main-table} summarizes our results. Overall, the $\mathcal{T}_{\mathrm{VAE}}$ has comparable performance when considering it for modelling i.i.d samples. In Appendix D we have a more extensive reporting of these results. There we can see that $\mathcal{T}_{\mathrm{VAE}}$ behaves differently than the two baselines: 
In general, it has a significantly better conditional log-likelihood (i.e. reconstruction error), but its latent distribution is farther from the employed prior than a VAE. 
We hypothesis that this phenomenon is due too the necessity to allocate additional information in latent space.

\begin{table}[t!]
\caption{Approximate log marginal likelihood on the MNIST (M), Fashion-MNIST (F) and Omniglot (O) test set using 1000 important samples from the posterior. We consider three different latent dimensions.
The column $r_\chi$ denotes the three different variants introduced earlier, Additive (A), Matrix (M) and Neural (N).}
\vskip 0.15in
\begin{center}
\begin{small}
\begin{sc}
\begin{tabular}{lccccc}
\toprule
Model & $r_{\chi}$ & zdim & \multicolumn{3}{c}{$\log p(\mathbf{x_0})$}  \\
\midrule
                             &      &      & M      & F       & O                 \\ 
\midrule
$\mathcal{T}_{\textrm{VAE}}$ & A    & 100  & -91.6  & - 236.0 & - 131.6 \\
$\mathcal{T}_{\textrm{VAE}}$ & M    & 100  & -92.4  & - 237.5 & - 130.4 \\
$\mathcal{T}_{\textrm{VAE}}$ & N    & 100  & -91.7  & - 237.9 & - 139.6 \\
\textrm{VAE+}             &     -   & 100  & -92.1  & - 236.3 & - 125.9 \\
\textrm{VAE}              &     -   & 100  & -90.0  & - 234.2 & - 128.9 \\
\midrule
$\mathcal{T}_{\textrm{VAE}}$ & A    & 25    & -91.2 & - 235.3 & - 127.9 \\
$\mathcal{T}_{\textrm{VAE}}$ & M    & 26    & -90.9 & - 235.7 & - 127.3 \\
$\mathcal{T}_{\textrm{VAE}}$ & N    & 25    & -90.5 & - 234.8 & - 127.3 \\
\textrm{VAE+}            &      -   & 25    & -92.5 & - 236.9 & - 126.6 \\
\textrm{VAE}             &      -   & 25    & -91.0 & - 234.1 & - 130.1 \\
\midrule
$\mathcal{T}_{\textrm{VAE}}$ & A    & 10  & -94.1 & - 235.4 & - 134.0 \\
$\mathcal{T}_{\textrm{VAE}}$ & M    & 10  & -94.7 & - 235.6 & - 133.7 \\
$\mathcal{T}_{\textrm{VAE}}$ & N    & 10  & -93.5 & - 235.3 & - 134.1 \\
\textrm{VAE+}           &        -  & 10  & -95.4 & - 236.8 & - 131.7 \\
\textrm{VAE}               &     -  & 10  & -91.8 & - 235.1 & - 133.7 \\
\bottomrule
\end{tabular}
\label{tab:generative-main-table}
\end{sc}
\end{small}
\end{center}
\vskip -0.1in
\end{table}

\paragraph{Latent embeddings.}
The results for $\mathcal{T}_{\textrm{VAE}}$ with respect to conditional log-likelihoods (cf. previous paragraph) indicate that the latent representations from the encoder $q_\phi(z|x)$ are different from those attained from a standard VAE. More specifically, we believe that $\mathcal{T}_{\textrm{VAE}}$ learns a latent space where the neighborhood of a sample is populated and structured in a different way than in a VAE. We investigate this structure by considering a challenging \emph{out-of-sample classification task} that uses a trained encoder to produce representations for samples that should be classified by an additional downstream model.

As a classifier we use a simple k-nearest neighbour (KNN) classifier --- this is on purpose to avoid training another set of parameters. The \emph{out-of-sample} aspect is due to the way the encoder is trained and used: We train the encoder (by training $\mathcal{T}_{\textrm{VAE}}$, or the baselines, VAE and VAE+) on one dataset, but then use this encoder \emph{to embed a different dataset}. Of course, this can not work for arbitrary pairs of dataset. 
The datasets should be from the same modality (in our case, $28\times28$ grayscale images) and should have a similar level of complexity (in our case, depiction of one conceptual entity on a black background). The connecting element between the datasets with respect to the $\mathcal{T}_{\textrm{VAE}}$ model is that, because of the same modality, the datasets \emph{share the same set of applicable transformations}. Variations within a class are due to a set of (symmetric) transformations, so if latent embeddings are done in such a way that they adhere to these transformations, even out-of-sample data should have consistent (that is, concerning the class information at least) latent representations.

In the experiments reported in Table~\ref{tab:knn-main-table} we look at the following two settings: (i) Training an encoder on MNIST and applying a KNN classifier on out-of-sample encodings from the AffNIST dataset and (ii) training an encoder on Fashion-MNIST and applying a KNN classifier on out-of-sample encodings from AffNIST. 
More specifically, say for the first setting, we train a $\mathcal{T}_{\textrm{VAE}}$ (and the baselines) on MNIST. The resulting encoder is then used to map 100000 samples from the AffNIST training set and the full AffNIST test set (320000 samples) to latent representations. Latent representations are formed by taking the mean of the approximate Gaussian posterior\footnote{We also properly sampled from the approximate posterior to get latent embeddings, but this approach led to significantly worse results for the baseline models.}. The 100000 AffNIST embeddings from the training set then form the training data for the non-parametric KNN classifier, which is evaluated on the test set embeddings. We use $k=5$, the number of neighbours, in all experiments and the $L_2$ norm between embeddings as the distance metric.

\begin{table}[t!]
\caption{KNN classification accuracy for out-of-sample embeddings. We train encoders on MNIST (M) and Fashion-MNIST (F), and use these to embed training and test data from AffNIST. The column labelled \textsc{M-Aff} indicates that the encoder is trained on MNIST, but the encoded data for the actual classification task is from AffNIST, i.e. the KNN is trained and evaluated on these out-of-sample embeddings. The column \textsc{M} indicates that the encoder is trained on MNIST and the embedding (and hence KNN classification) is also done on MNIST (the same applies to column \textsc{F} accordingly).We show this setting here, too, in order to make sure that the baseline encoders don't suffer from underfitting. The improvement in performance is particularly remarkable when there is a large domain shift between training set (Fashion-MNIST) and test set (AffNIST). The column $r_\chi$ denotes the three different variants introduced earlier, Additive (A), Matrix (M) and Neural (N).}
\vskip 0.15in
\begin{center}
\begin{small}
\begin{sc}
\begin{tabular}{lcccccc}
    \toprule
    Model & $r_{\chi}$ & zdim  & M & M-Aff & F & F-Aff                              \\ 
    \midrule
    $\mathcal{T}_{\textrm{VAE}}$ & A  & 100 & 0.98  & \textbf{0.85}  & 0.87  & 0.78 \\
    $\mathcal{T}_{\textrm{VAE}}$ & M  & 100 & 0.98  & 0.83  & 0.86  & \textbf{0.80}\\
    $\mathcal{T}_{\textrm{VAE}}$ & N  & 100 & 0.98  & 0.81           & 0.86  & 0.73\\
    VAE+                         & -  & 100 & 0.98  & 0.75           & 0.84  & 0.44 \\ 
    VAE                          & -  & 100 & 0.96  & 0.64           & 0.84  & 0.33\\  
    \midrule
    $\mathcal{T}_{\textrm{VAE}}$ & A  & 25  & 0.98 & 0.82          & 0.86  & 0.73\\ 
    $\mathcal{T}_{\textrm{VAE}}$ & M  & 26  & 0.98 & \textbf{0.84} & 0.86  & \textbf{0.74}\\ 
    $\mathcal{T}_{\textrm{VAE}}$ & N  & 25  & 0.98 & 0.81          & 0.86  & 0.69\\ 
    VAE+        &               -     & 25  & 0.98 & 0.76          & 0.84  & 0.41\\  
    VAE         &               -     & 25  & 0.96 & 0.58          & 0.83  & 0.34\\ 
    \midrule
    $\mathcal{T}_{\textrm{VAE}}$ & A  & 10  & 0.97  & 0.62          & 0.84 &  0.48 \\
    $\mathcal{T}_{\textrm{VAE}}$ & M  & 10  & 0.97  & 0.63          & 0.84 &  0.48\\
    $\mathcal{T}_{\textrm{VAE}}$ & N  & 10  & 0.97  & 0.59          & 0.84 &  \textbf{0.49}\\
    VAE+     &                  -     & 10  & 0.97  & \textbf{0.65} & 0.83 &  0.41\\  
    VAE      &                  -     & 10  & 0.96  & 0.57          & 0.83 &  0.31\\ 
    \bottomrule
    \end{tabular}
\label{tab:knn-main-table}
\end{sc}
\end{small}
\end{center}
\vskip -0.1in
\end{table}

Overall, we don't expect that the baselines can perform too well in this domain shift scenario --- nothing in their architecture supports this task explicitly. Because MNIST and AffNIST are relatively similar with respect to their content we would expect that at least VAE+ should perform acceptable in the MNIST/AffNIST setting: VAE+ is trained, due to the data augmentation, on samples that are very similar to samples from AffNIST. In the case of F-MNIST/AffNIST however the structured latent space of a $\mathcal{T}_{\textrm{VAE}}$ should be much more helpful for out-of-sample embeddings. 

In general Table~\ref{tab:knn-main-table} shows that $\mathcal{T}_{\textrm{VAE}}$ performs significantly better. In order to ensure that the trained encoders on the \emph{base datasets} are not suffering from underfitting for VAE/VAE+, we also run the classification experiments without any domain shift (i.e. we use MNIST to train the encoder, and then classify embeddings from the MNIST test set using embeddings from the MNIST training set). Because the test sets are supposedly in-distribution, we would not expect any differences in the classification results for all models, which is also reflected in Table~\ref{tab:knn-main-table}. 

\begin{figure*}[t!]
\begin{center}
\subfigure[\textsc{M-Aff}]{
\includegraphics[scale=0.45]{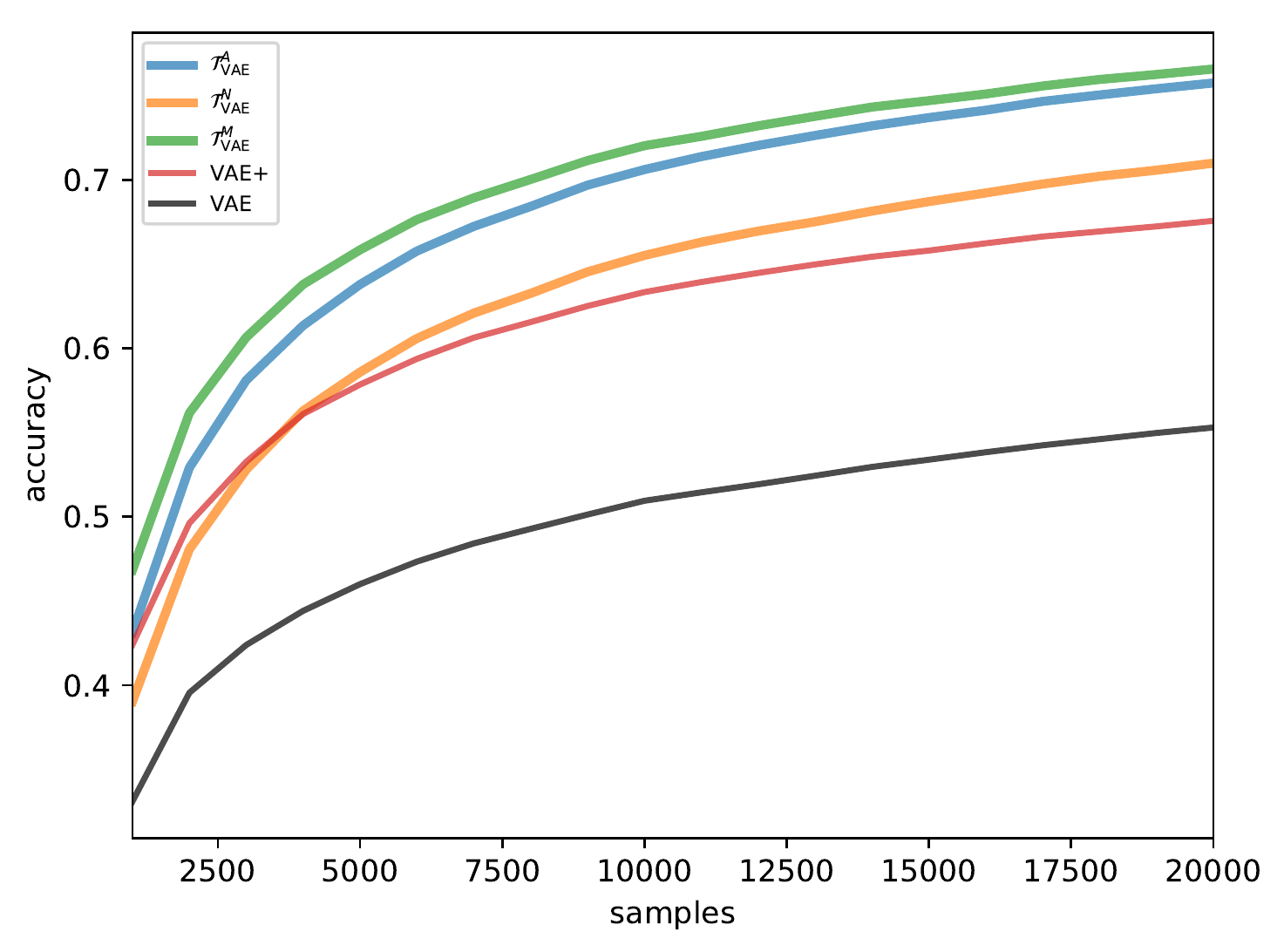}
}
\quad\quad\quad
\subfigure[\textsc{F-Aff}]{
\includegraphics[scale=0.45]{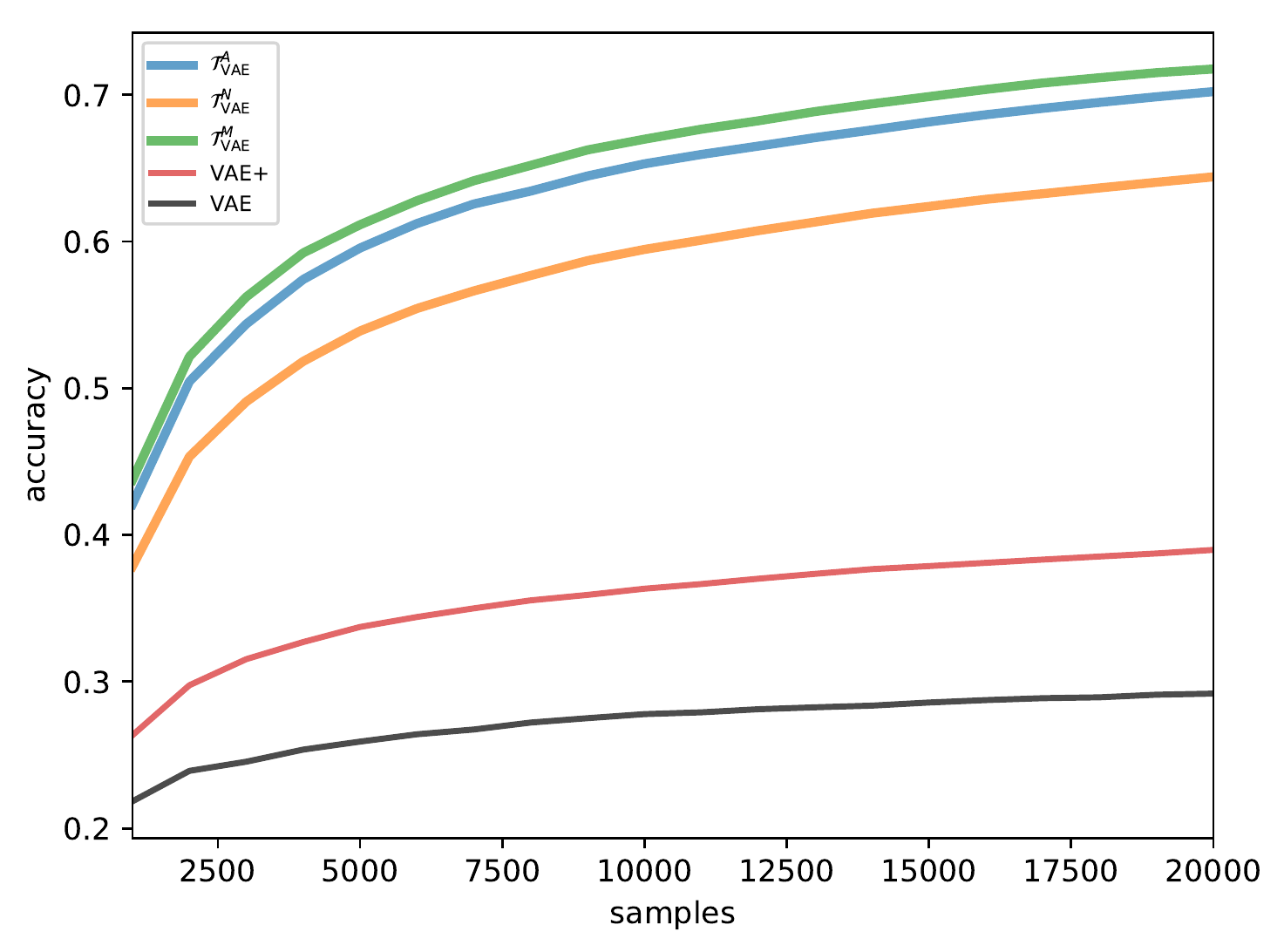}
}
\caption{Improved sample efficiency for $\mathcal{T}_{\textrm{VAE}}$ in out-of-sample classification tasks. The setting is the same as in Table~\ref{tab:knn-main-table}, encoders are trained on a dataset that is different from the dataset that is actually encoded for the KNN classification task. Here we are interested in \emph{the rate of improvement of the classification accuracy} while increasing the amount training samples \emph{for the KNN classifier}.  a) Classification accuracy curves for the \textsc{M-Aff} setting, i.e. encoders are trained on MNIST, samples from AffNIST are encoded and classified. We use at most 20\% of the available samples from the AffNIST training set. b) Classification accuracy curves for the \textsc{F-Aff} setting. Again the desired effect is most pronounced in the setting where the domain shift is more extreme. KNN settings are the same as in Table~\ref{tab:knn-main-table}.
}
\label{fig:sample_efficiency}
\end{center}
\vskip -0.2in
\end{figure*}

We also hypothesize that in the setting of a domain shift (e.g. embedding trained on F-MNIST, used for AffNIST), the latent structure induced by $\mathcal{T}_{\textrm{VAE}}$ should also result in a better sample efficiency with respect to the size of the training set for the classification task. That is, increasing the training set should lead to larger gains in the classification accuracy for the embeddings derived from $\mathcal{T}_{\textrm{VAE}}$. Obviously this will be most notably in the small data regime. 
This hypothesis is corroborated by the experiments summarized in Figure~\ref{fig:sample_efficiency} for the two settings described previously. The improved sample efficiency is reflected by the steeper gradient of the various classification accuracy curves for $\mathcal{T}_{\textrm{VAE}}$ models. The effect is most pronounced for the very challenging F-MNIST/AffNIST setting.

\section{Related Work} \label{sec:related}

{\bf Representing Transformations.}
Encoding physical and mathematical properties \citep{wood1996representation, bloem2019probabilistic} for informing statistical learning algorithms is a long-standing challenge in machine learning.
When some properties of the input data are known, many approaches focus on building invariance and equivariance directly into the architecture \citep{cohen2016group, lenssen2018group}. Our work continues the line of research that looks into encoding transformations explicitly or implicitly into latent variable models \citep{memisevic2007unsupervised, memisevic2012multi, michalski2014modeling, cohen2014transformation, schweizer2011probabilistic, falorsi2018explorations}. Differently from these approaches we integrate the idea of latent transformations into the framework of VAEs by building hierarchical models that are informed by desirable algebraic constraints of the resulting model. From a more abstract point of view, we are trying to learn commutativity between spaces with the goal of building representations that can deal with variability, distortions and viewpoint change \citep{achille2018emergence} given a generic approximate group structure of the transformations of a domain. \citep{poggio2015theory, Poggio:2013, anselmi2014representation}. 

{\bf Causality.} The idea that objects (and their representations) must be defined by their transformation properties was our starting point, which we approached from the perspective of bringing together representation theory (in the context of group theory) and representation learning (in the context of latent variable models). However, this very perspective has also been motivated from the angle of \emph{causality}~\citep{scholkopf2019causality}. In particular, we should highlight the paper~\citep{parascandolo2017learning}, where they approached the problem of data going under transformations with \emph{ the principle of independent mechanisms}~\citep{peters2017elements}. It is not clear how one can tie an approach that is motivated in group theory directly to the approach rooted in causality and the independent mechanisms assumption, but we leave exploring possible connections for future research.

{\bf Disentanglement.} 
Borrowing ideas from physics, a disentangled representation can be defined as the one that preserves the largest amount of invariances in the data~\cite{higgins2018towards}.
Different approaches regarding the question of \emph{disentangled representations} focus on different properties of desirable representations, for example predictability (or rather \emph{un}-predictability) \cite{schmidhuber1992learning,brakel2017learning} or encouraging sparsity in the representation~\cite{olshausen2004sparse} or maximizing statistical measures of information~\cite{kim2018disentangling, chen2018isolating}, or exploiting symmetries~\cite{caselles2019symmetry} or explicitly accounting for transformations~\cite{detlefsen2019explicit}.
However, there is evidence that a disentangled representation is neither feasible without additional assumptions \cite{locatello2018challenging, shu2019weakly}, nor is the underlying model identifiable \cite{khemakhem2019variational, hauberg2018only}. Here, we did not aim to learn a disentangled representation, but to learn a model for data and their transformations. These goals seem intuitively related but grounding their relations is beyond the scope of this work.

{\bf Conditional Latent Variable Models.} Having access to a context (i.e. prior information) improves the capacity of latent variable models.
\citep{eslami2018neural, kumar2018consistent} introduced a generative rendering engine that conditions on camera viewpoints and can be queried to generate new views.
Instead of camera coordinates, we use transformation \emph{properties} as anchors for our model.
\citep{graves2018associative} aim to improve the representation for i.i.d. data in the presence of powerful AR decoders, learning a prior over $z$ conditioned on the top KNN in latent space associated with a given sample point.  Our approach is orthogonal: we similarly break the i.i.d. structure but we resolve to learn transformations between samples.
In general our model can be improved by additional prior information as investigated in other work, for example learned priors can be used for improving the model's generative capacity \citep{tomczak2017vae}, for meta-learning strategies \citep{edwards2016towards, snell2017prototypical, garnelo2018neural}, or imposing invariance properties \citep{nalisnick2018learning}.

\section{Conclusion}
Objects are defined by their behaviour under transformations \citep{klein}. This very simple but also very fundamental assumption is the starting point  of our investigations in this paper. From the perspective of representation learning this means that representations for observed entities can not be modeled or learned in isolation but need to be put in relation to the transformations of their domain.

In this work we approach this insight by introducing the concept of a latent transformation into the framework of variational autoencoders \citep{kingma2013auto, rezende2014stochastic}. Hereby we continue an exciting line of previous research directions \citep{memisevic2012multi, michalski2014modeling, cohen2014transformation} and merge these with ideas of amortized variational inference \citep{gershman2014amortized}.

Our main conceptual contribution is to construct a novel hierarchical graphical model, the $\mathcal{T}_{\textrm{VAE}}$, according to properties of an idealized group structure, which enables the learning of latent representations of data as well as of transformations at the same time. We show how this model can be efficiently trained through a learning objective based on the variational lower bound that we drive. 
In qualitative experiments we show that the inferred latent transformations contain semantically reasonable information and behave accordingly to the intended group structure. In quantitative experiments we demonstrate that for a challenging out-of-sample task the latent embeddings induced by the $\mathcal{T}_{\textrm{VAE}}$ outperform reasonable baselines significantly --- generalization to out-of-sample data is one of the core challenges in Machine Learning and we believe that at least our conceptual approach is a promising path for finding solutions for this challenge.

An important next step is to enhance the proposed approach in such a way that it no longer is necessary to define the utilized transformations by hand. This means specifically that we need to apply our approach to time-series data. 
For this type of data, the assumption of an invertible transformation that connects three consecutive time steps is probably too constrained. 
So a concrete goal in our line of research is to identify alternative algebraic (i.e. structural) properties that encode semantic properties of time-series data \citep{gregor2018temporal}. 
More generally, we are interested to model non-bijective transformations and express these in latent space (for example the algorithmic mapping of an image to its segmented instance). We believe that in order to make strides forward in these questions, it is important to understand how to encode algebraic equivalence relationships in graphical models. One important direction of research to tackle this question is to utilize directed and undirected graphical models within the framework of amortized variational inference \emph{at the same time} \citep{kuleshov2017neural}.

\clearpage
\section*{Acknowledgement}
We would like to thank 
Justin Bayer,
Pierluca D'Oro, 
Marco Gallieri,
Jan Eric Lenssen, 
Simone Pozzoli, 
Pranav Shyam, 
Jerry Swan, 
Timon Willi 
for insightful comments and discussions.

\bibliography{biblio.bib}
\bibliographystyle{util/icml2020}

\newpage
\onecolumn
\appendix
\section{Model}
In this section we describe the model introduced in Section 2 of the main text in more detail. We derive the lower bound step by step and state the actual training objective after applying the changes mentioned in Section 3.2 of the main text to the lower bound.

\begin{figure}[b]
\begin{center}
\includegraphics[scale=0.7]{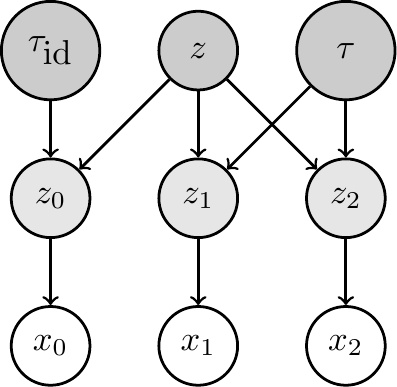}
\caption{The graphical model for the transformation-aware VAE, $\mathcal{T}_{\mathrm{VAE}}$}.
\label{fig:app:tauvae}
\end{center}
\end{figure}

\subsection{Generative Model}
The graphical model of a $\mathcal{T}_{\mathrm{VAE}}$ is shown in Figure~\ref{fig:app:tauvae}. We factorize the (general) joint distribution $p((x_0, x_1, x_2), (z_0, z_1, z_2), z, \tau, \tau_{\mathrm{id}})$ in a very straight-forward way (introducing parameters $\theta$ and $\chi$):
\begin{equation}
p_{\theta}(x_0 \vert z_0)
p_{\theta}(x_1 \vert z_1)
p_{\theta}(x_2 \vert z_2)
r_{\chi}(z_0 \vert z,\tau_{\mathrm{id}})
r_{\chi}(z_1 \vert z, \tau)
r_{\chi}(z_2 \vert z, \tau^{-1})
\pi(z)
\pi(\tau)
\delta(\tau_{\textrm{id}})
\label{eq:app:genmod}
\end{equation}
where $\pi(\cdot)$ refers to some prior for now and the prior $\delta(\tau_{\mathrm{id}})$ on the random variable $\tau_{\rm id}$ is considered to be a delta function centered at the identity of the transformation group.

\subsection{ELBO}
\label{app:elbo}
We learn the generative model from Eq.~\ref{eq:app:genmod} by maximizing the triplet log-likelihood. Because this is in general intractable, we have to resort to identifying a lower-bound on this log-likelihood and optimize it instead. The lower-bound is derived by introducing an \emph{approximate} posterior distribution $q(\tau_{\textrm{id}}, \tau, z, (z_0, z_1, z_2) \vert (x_0, x_1, x_2))$ and applying Jensen's inequality to the joint $\log p((x_0, x_1, x_2))$ like so:
\begin{align}
 \log &\int  p((x_0, x_1, x_2), (z_0, z_1, z_2), \tau, \tau_{\textrm{id}} z) dz dz_0 dz_1 dz_2 d\tau d \tau_{\textrm{id}} \geq \\
      &\int q(\tau_{\textrm{id}}, \tau, z, (z_0, z_1, z_2) \vert (x_0, x_1, x_2)) \log \frac{p((x_0, x_1, x_2), (z_0, z_1, z_2), \tau, \tau_{\textrm{id}}, z)}{q(\tau_{\textrm{id}}, \tau, z, (z_0, z_1, z_2) \vert (x_0, x_1, x_2))} dz dz_0 dz_1 dz_2 d \tau d\tau_{\textrm{id}}
      \label{eq:app:lb1}
\end{align}

Tractability of this lower bound depends to a large degree on how the approximate posterior is decomposed. In this work, we chose the following simple one:
\begin{equation}
q_{\xi}(\tau \vert z_1, z_2) 
q_{\psi}(z \vert z_1, z_2)
q_{\phi}(z_0 \vert x_0)
q_{\phi}(z_1 \vert x_1)
q_{\phi}(z_2 \vert x_2).
\label{eq:app:post}
\end{equation}
Different decompositions are left for future work. Importantly, we set $q(\tau_{\textrm{id}}|\tau, z, z_0, z_1, z_2) \equiv 1$. Also note that in the decomposition above we decided to ignore $z_0$ and $z$  for $q_{\xi}(\tau|z_1, z_2 )$ and $z_0$ for $q_\psi(z \vert z_1, z_2)$ by choice.
Thus, for the lower bound in Eq.~\ref{eq:app:lb1} we get

\begin{equation}
\int 
q_{\xi}(\tau \vert z_1, z_2)
q_{\psi}(z \vert z_1, z_2)
\prod_{i=0}^2
q_{\phi}(z_i \vert x_i) 
\log \frac{p((x_0, x_1, x_2), (z_0, z_1, z_2) \vert \tau, \tau_{\textrm{id}}, z) \pi(z) \pi(\tau)}
{
q_{\xi}(\tau \vert z_1, z_2)
q_{\psi}(z \vert z_1, z_2)
\prod_{i=0}^2
q_{\phi}(z_i \vert x_i)
} 
dz dz_0 dz_1 dz_2 d\tau.
\end{equation}

This can be written more compactly, using the shorthand $\mathbf{q}_{\xi \psi \phi} \equiv
q_{\xi}(\tau \vert z_1, z_2)
q_{\psi}(z \vert z_1, z_2)
q_{\phi}(z_1 \vert x_1) 
q_{\phi}(z_2 \vert x_2)$, as

\begin{align}
\begin{split}
\mathbb{E}_{\mathbf{q}_{\xi \psi \phi}q_{\phi}(z_0 \vert x_0)}
[
&  \log p_{\theta}(x_2 \vert z_2) + 
  \log p_{\theta}(x_1 \vert z_1) + 
  \log p_{\theta}(x_0 \vert z_0) \\
-& \log q_{\phi}(z_2 \vert x_2)
- \log q_{\phi}(z_1 \vert x_1)
- \log q_{\phi}(z_0 \vert x)
- \log q_{\psi}(z \vert z_1, z_2)
- \log q_{\xi}(\tau \vert z_1, z_2)\\
+& \log r_{\chi}(z_2 \vert z,\tau^{-1}) +
  \log r_{\chi}(z_1 \vert z,\tau) + 
  \log r_{\chi}(z_0 \vert z,\tau_{\mathrm{id}}) + 
  \log \pi(z) + \log \pi(\tau)
].
\end{split}
\end{align}

Reorganizing terms leads to 
\begin{align}
\begin{split}
&\mathbb{E}_{q_{\phi}(z_0 \vert x_0)}
\left[
     \log p_{\theta}(x_0 \vert z_0)\right] - 
     \mathbb{E}_{\mathbf{q}_{\xi \psi \phi}q_{\phi}(z_0 \vert x_0)}
     \left[
      \log q_{\phi}(z_0 \vert x_0) 
      - \log r_{\chi}(z_0 \vert z,\tau_{\mathrm{id}})
      \right] \\
+& \mathbb{E}_{q_{\phi}(z_1 \vert x_1)}
\left[
     \log p_{\theta}(x_1 \vert z_1)\right] - 
     \mathbb{E}_{\mathbf{q}_{\xi \psi \phi}}
     \left[
      \log q_{\phi}(z_1 \vert x_1) 
      - \log r_{\chi}(z_1 \vert z,\tau)
      \right] \\
+& \mathbb{E}_{q_{\phi}(z_2 \vert x_2)}
\left[
     \log p_{\theta}(x_2 \vert z_2)\right] - 
     \mathbb{E}_{\mathbf{q}_{\xi \psi \phi}}
     \left[
      \log q_{\phi}(z_2 \vert x_2) 
      - \log r_{\chi}(z_2 \vert z,\tau^{-1})
      \right] \\
-& \mathbb{E}_{\mathbf{q}_{\xi \psi \phi}}
\left[
    \log q_{\psi}(z \vert z_1, z_2) - \log \pi(z)
\right]
- \mathbb{E}_{\mathbf{q}_{\xi \psi \phi}}
\left[
    \log q_{\xi}(\tau \vert z_1, z_2) - \log \pi(\tau)
\right],
\end{split}
\end{align}

and finally we obtain
\begin{align}
\begin{split}
\log p((x_0,  x_1, x_2))
\geq &\mathbb{E}_{q_{\phi}(z_0 \vert x_0)}
     \log p_{\theta}(x_0 \vert z_0)
     - \mathbb{E}_{q_{\phi}(z_0 \vert x_0)}
     \mathcal{D}_{\tau}\left[q_{\phi}(z_0 \vert x_0), r_{\chi}(z_0 \vert z,\tau_{\mathrm{id}}) \right] \\
+& \mathbb{E}_{q_{\phi}(z_1 \vert x_1)}
     \log p_{\theta}(x_1 \vert z_1)
     - \mathcal{D}_{\tau}\left[q_{\phi}(z_1 \vert x_1), r_{\chi}(z_1 \vert z,\tau) \right] \\
+& \mathbb{E}_{q_{\phi}(z_2 \vert x_2)}
     \log p_{\theta}(x_2 \vert z_2)
     - \mathcal{D}_{\tau}\left[q_{\phi}(z_2 \vert x_2), r_{\chi}(z_2 \vert z,\tau^{-1}) \right]  \\
-& \mathcal{D}_{\tau}
    \left[ q_{\psi}(z \vert z_1, z_2),  \pi(z)\right]
- \mathcal{D}_{\tau}
    \left[ q_{\xi}(\tau \vert z_1, z_2),  \pi(\tau)
    \right].
    \label{eq:app:finalelbo}
\end{split}
\end{align}

where 
$\mathcal{D}_{\tau}\left[x, y\right] \equiv
\mathbb{E}_{
    q_{\xi}(\tau \vert z_1, z_2)
    q_{\psi}(z \vert z_1, z_2)
    q_{\phi}(z_1 \vert x_1) 
    q_{\phi}(z_2 \vert x_2)
    } \log x / y
$.

What type of distribution should $q_\xi(\tau | z_1, z_2)$ be? Clearly it is tied to the way the random variable $\tau$ is used in $r_\chi(z_i \vert \cdot)$. 
For the implementation presented in this paper, $\tau$ is computed deterministically 
\footnote{Another simple way to parameterize $\tau$ is using a flow-based model and considering a different approximate posterior.}
\begin{equation} \label{eq:appendix:tau}
q_{\xi}(\tau \vert z_1, z_2) = \delta(\tau - f_\xi(z_1, z_2)),
\end{equation} 
where $z_1$ and $z_2$ are sampled from $q_\phi(z_1 \vert x_1)$ and $q_\phi(z_2 \vert x_2)$ respectively. 
$\tau_{\textrm{id}}$ is fixed to be the correct identity element, depending on the form or $r_\chi(\cdot)$. 


Considering the deterministic approximation for $\tau$, $\mathcal{D}_{\tau}$ simplifies to:
\begin{equation}
\mathcal{D}\left[x, y\right] \equiv
\mathbb{E}_{
    q_{\psi}(z \vert z_1, z_2)
    q_{\phi}(z_1 \vert x_1) 
    q_{\phi}(z_2 \vert x_2)
    } \log x / y.
\label{eq:app:div}
\end{equation}
In addition, since $\tau$ is computed deterministically (Eq.~\ref{eq:appendix:tau}), there will not be any learning signal coming from the divergence term $\mathcal{D}_{\tau}
    \left[ q_{\xi}(\tau \vert z_1, z_2),  \pi(\tau)
    \right]$ and it is dropped. Putting all together, 
    we arrive at the learning objective presented in Eq. \ref{eq:lb} in the main text.

\subsection{Training loss}
As we point out in the main paper, some adaptations of the derived lower bound in Eq.~\ref{eq:app:finalelbo} are necessary:
\begin{itemize}
\item Terms $\mathcal{D}\left[x, y\right]$ involving $r_\chi$  are substituted by Maximum Mean Discrepancy using a linear kernel (Eq.~\ref{eq:mmd}).
\item Empirically, we found out that the inferred latent variable $z$ does not reconstruct well it's associated observation $x_0$, i.e. $p_\theta(x_0 \vert z)$ has low likelihood. We add this as an additional term, substituting the expression $\mathcal{D}\left[q_{\phi}(z_0 \vert x_0), r_{\chi}(z_0 \vert z, \tau_{\mathrm{id}}) \right]$ with Eq.~\ref{eq:constraint}.
\end{itemize}

The overall training objective for $\mathcal{T}_{\mathrm{VAE}}$ is 
\begin{equation}
 \mathcal{L}(\theta, \phi, \psi, \chi, \xi) = 
\mathcal{L}_{D} + \mathcal{L}_{C} + \mathcal{L}_{x \vert z},
\end{equation}

where
\begin{equation}
\mathcal{L}_{D} = -\mathbb{E}_{q_{\phi}(z_0 \vert x_0)}
     \left[
     \log p_{\theta}(x_0 \vert z_0)
     \right]
     -\mathbb{E}_{q_{\phi}(z_1 \vert x_1)}
     \left[
     \log p_{\theta}(x_1 \vert z_1)
     \right]
     -\mathbb{E}_{q_{\phi}(z_2 \vert x_2)}
     \left[
     \log p_{\theta}(x_2 \vert z_2)
     \right],
\end{equation}

\begin{equation}
\mathcal{L}_{C} =
\left \Vert  \mathbb{E}_{q_\phi(z_1 \vert x_1)}[z_1] - \mathbb{E}_{r_\chi(z_1 \vert z, \tau)}[ z_1] \right \Vert^2
+
\left \Vert  \mathbb{E}_{q_\phi(z_2 \vert x_2)}[z_2] - \mathbb{E}_{r_\chi(z_2 \vert z, \tau^{-1})}[ z_2] \right \Vert^2
+
\mathcal{D}_{\textrm{KL}}\left[q_{\phi}(z_0 \vert x_0), \pi(z_0) \right],
\end{equation}

\begin{equation}
\mathcal{L}_{x \vert z} = -\mathbb{E}_{q_{\psi}(z \vert z_1, z_2)q_{\phi}(z_1 \vert x_1)q_{\phi}(z_2 \vert x_2)} 
    \left[\log p_{\theta}(x_0 \vert z)\right] + \mathcal{D}
    \left[ q_{\psi}(z \vert z_1, z_2),  \pi(z)
    \right]
\end{equation}
Note the sign change, as we train by \emph{minimizing} $\mathcal{L}(\theta, \phi, \psi, \chi, \xi)$. 
We use the standard pathwise derivative estimator (i.e. reparamterization trick \citep{kingma2013auto, schulman2015gradient}) when computing gradients where sampling operations are involved. Variables $z$, $z_0$, $z_1$ and $z_2$ in the above losses are in general inferred according to the approximate posterior in Eq.~\ref{eq:app:post}.
\clearpage
\section{Algorithm}
\label{appendix:algorithm}

\begin{algorithm}
   \caption{$\mathcal{T}_{\mathrm{VAE}}$}
   \label{alg:example}
\begin{algorithmic}
   \STATE {\bfseries Input:} $\mathcal{X}, \mathcal{F}$
   \REPEAT
   \STATE Sample  $x_0$, $f$
   \STATE Compute $x_2  = f \circ x$,
                  $x_1  = f^{-1} \circ x$,
    \STATE Encode $(z_0, z_1, z_2) \sim q_{\phi}(z \vert x)$ \\
    $\mathcal{L}_0 = \mathbb{E}_{q_{\phi}(z_0 \vert x_0)} \left[- \log p_{\theta}(x_0 \vert z_0) \right] + \mathcal{D}_{\textrm{KL}} \left[q_{\phi}(z_0 \vert x_0), \pi(z_0) \right]$ \\
                  $\mathcal{L}_1 = \mathbb{E}_{q_{\phi}(z_1 \vert x_1)} \left[- \log p_{\theta}(x_1 \vert z_1) \right]$                                                                  \\
                  $\mathcal{L}_2 = \mathbb{E}_{q_{\phi}(z_2 \vert x_2)} \left[- \log p_{\theta}(x_2 \vert z_2) \right]$                                                                  \\
                  $\mathcal{L}_{D}(\theta, \phi)= \mathcal{L}_0 + \mathcal{L}_1 + \mathcal{L}_2$          \\
    \BlankLine
            Sample $z \sim q_{\psi}(z \vert z_1, z_2 )$      \\
            Compute $\tau = \tau_{\xi}(z_1, z_2)$                                      \\
            Compute 
            $\tilde z_1 = r_{\chi}(z_1 \vert z, \tau)$, 
            $\tilde z_2 = r_{\chi}(z_2 \vert z, \tau^{-1})$                                    \\
            $\mathcal{L}_{C}(\phi, \psi, \chi) = \vert\vert \tilde z_1 - z_1 \vert\vert_{2} 
            + \vert\vert \tilde z_2 - z_2 \vert\vert_{2}$ \\
    \BlankLine
      $\mathcal{L}_{x \vert z}(\phi, \theta, \psi, \xi) = 
      \mathbb{E}_{q_{\psi}(z \vert z_1, z_2) 
                  q_{\phi}(z_1 \vert x_1) 
                  q_{\phi}(z_2 \vert x_2)} 
      \left[- \log p_{\theta}(x_0 \vert z) \right] + \mathcal{D} \left[q_{\psi}(z \vert z_1, z_2), \pi(z) \right]$ \\
    \BlankLine
      $\mathcal{L}(\theta, \phi, \psi, \chi, \xi) = 
        \mathcal{L}_{D} + \mathcal{L}_{C} + \mathcal{L}_{x \vert z}$  \\
      $(\theta, \phi, \psi, \chi, \xi) \leftarrow \nabla \mathcal{L}$                                       \\
    \BlankLine
    \label{algo:inVAE}
   \UNTIL{Convergence}
\end{algorithmic}
\end{algorithm}

\clearpage
\section{Observations}
\label{appendix:properties}
In this section we collect a set of experiments that we conducted so far with the goal to understand our proposed model better.

\begin{figure}[t]
\begin{center}
\subfigure[\textsc{M-Aff}]{
\includegraphics[scale=0.45]{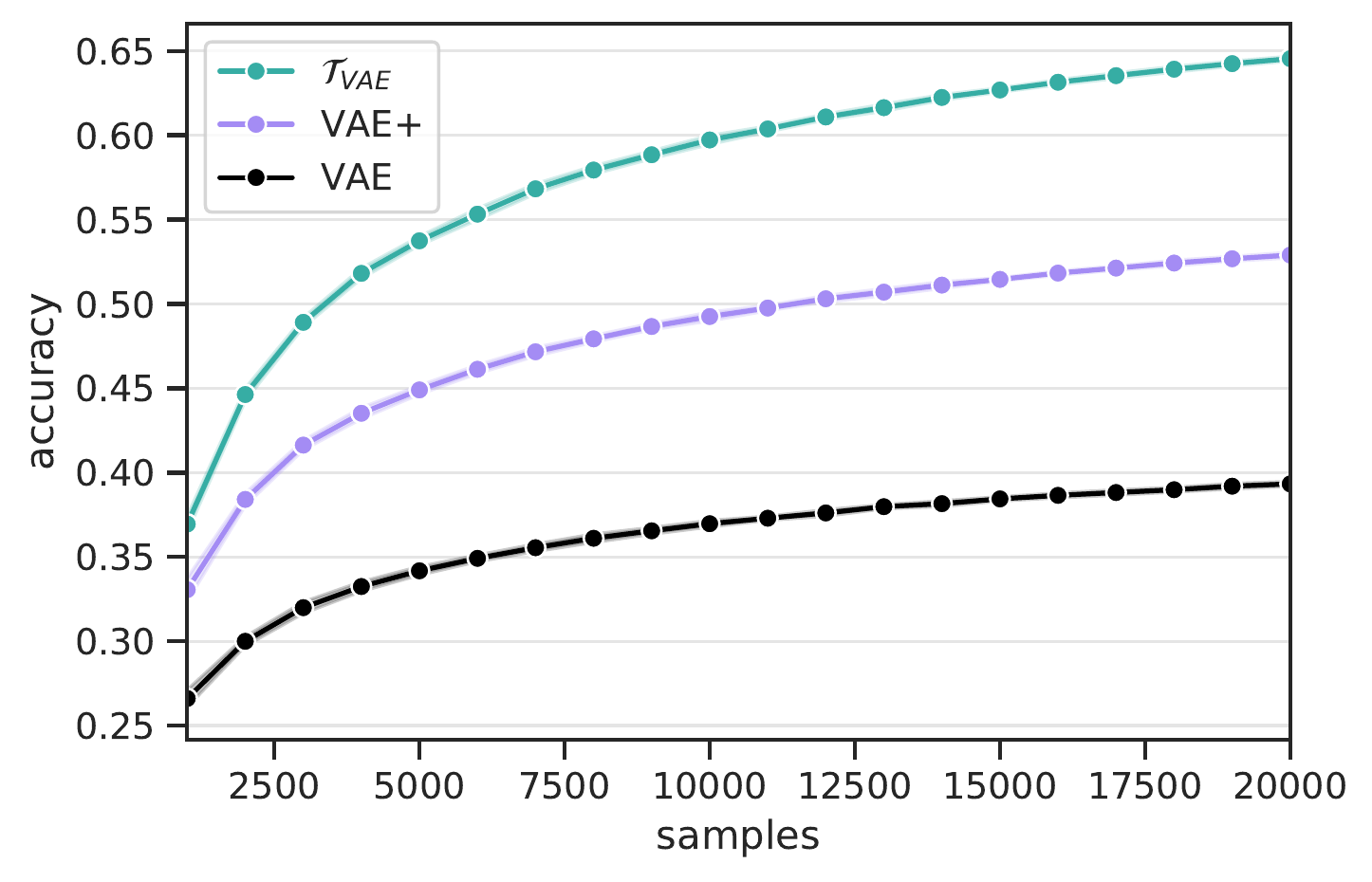}
}
\quad\quad\quad
\subfigure[\textsc{F-Aff}]{
\includegraphics[scale=0.45]{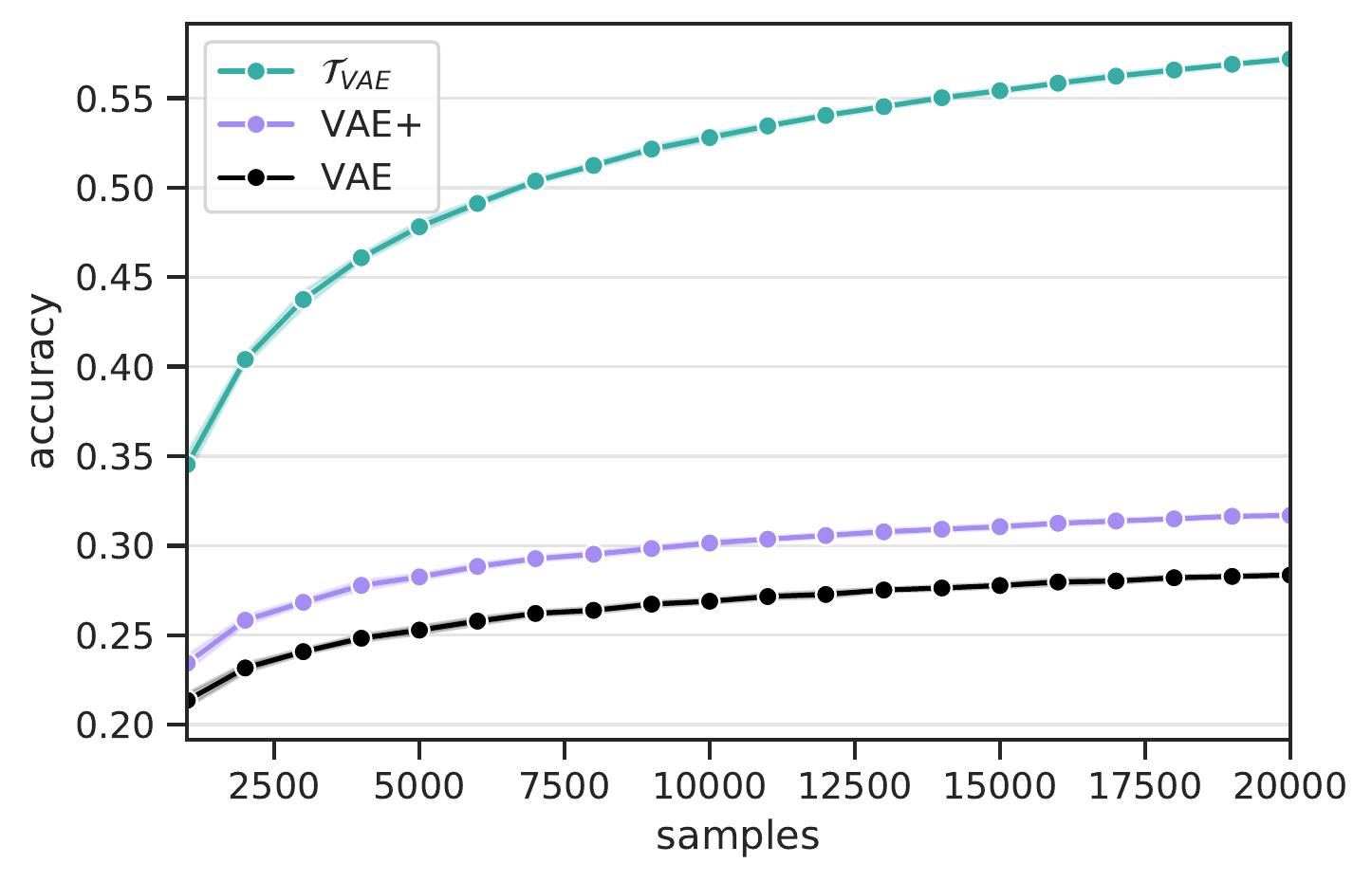}
}
\caption{(a) Classification accuracy on AffNIST test set over 20 runs using $\mu_{z}$. Models trained on MNIST. (b) Classification accuracy on AffNIST test set over 20 runs using $\mu_{z}$. Models trained on Fashion-MNIST. Best viewed electronically.}
\label{fig:app:sample_efficiency}
\end{center}
\end{figure}

\paragraph{Out-of-sample KNN experiments.} In Figure \ref{fig:app:sample_efficiency}(a) and (b) we report the same experiment as Figure 4 in the main paper. This time, we repeated this experiment twenty times, while varying the training set subsampling. The Figure shows the average behaviour and also confidence intervals for these twenty runs (the confidence intervals are very small, it is necessary to zoom in substantially to see this detail). We can see that the result is extremely stable and it is not dependent on the particular split chosen.

\begin{figure}[!h]
\begin{center}
\includegraphics[scale=0.45]{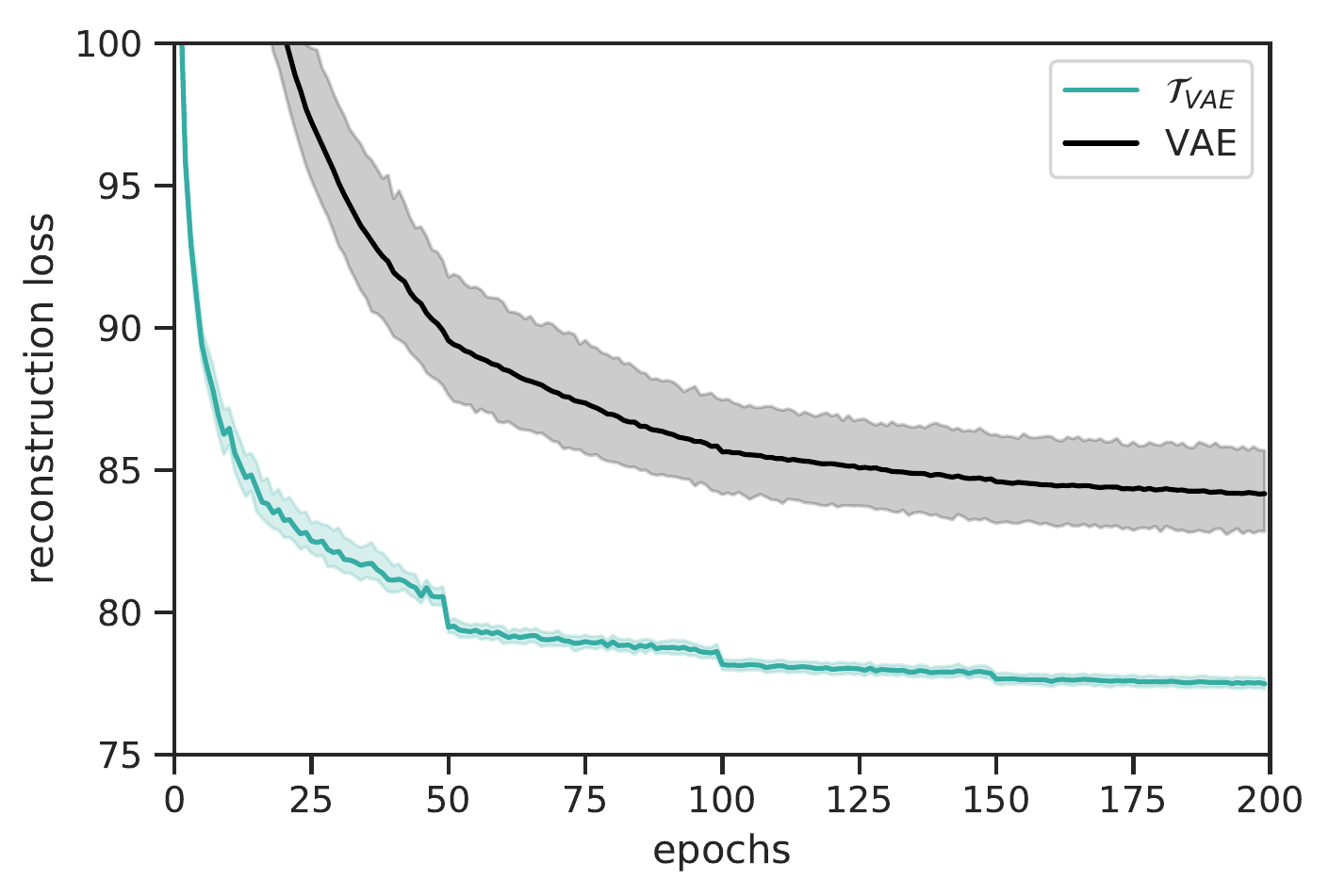}
\caption{Stability during training on MNIST with 10 random seeds, evaluated on MNIST test set.}
\label{fig:app:stab}
\end{center}
\end{figure}

\paragraph{Stability during training.} An interesting phenomenon is shown in Figure~\ref{fig:app:stab}. As an indicator for stability of training, we visualize the reconstruction loss for samples $x_0$ from the respective test set (in this case, MNIST). For the  $\mathcal{T}_{\mathrm{VAE}}$ we see a significant reduction in noisy behaviour compared to a standard VAE. This observation holds true for all variations of $r_\chi$ and all datasets used in the paper.

\begin{figure}
\vskip 0.2in
\begin{center}
\subfigure[$\mathcal{T}_{\textrm{VAE}} vs. \textrm{VAE}$]{
\includegraphics[width=.45\linewidth]{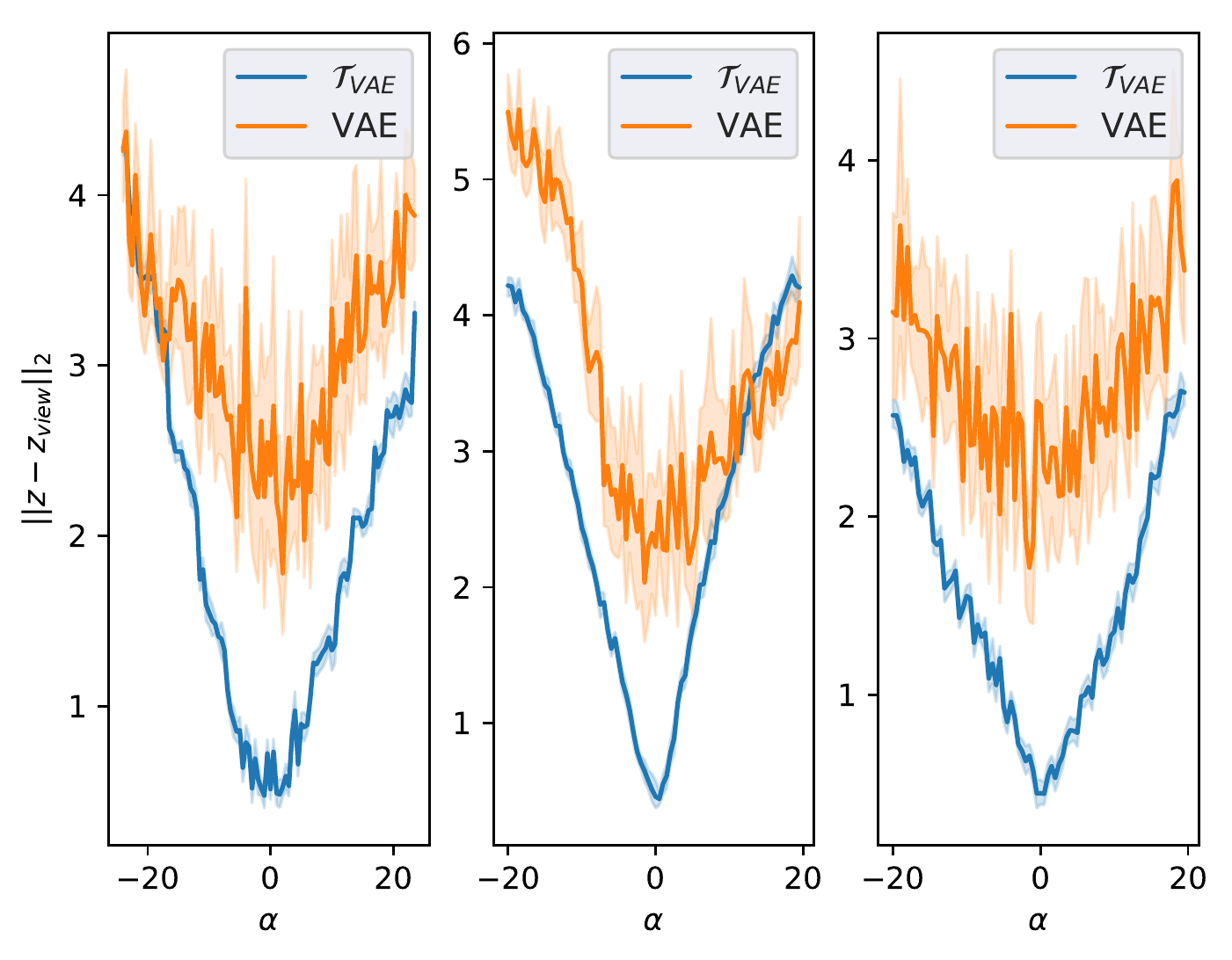}
}
\subfigure[$\mathcal{T}_{\textrm{VAE}} vs. \textrm{VAE+}$]{
\includegraphics[width=.45\linewidth]{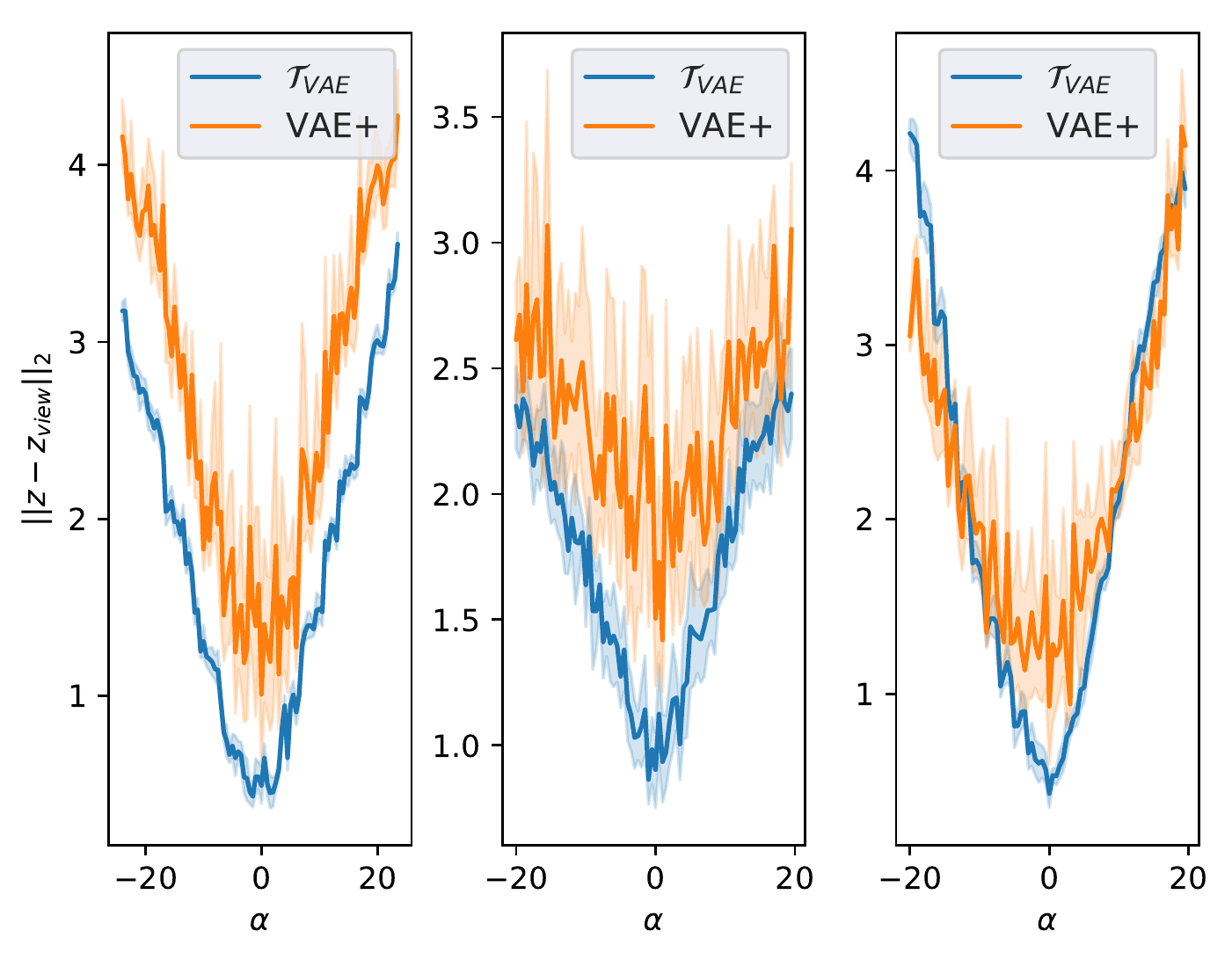}
}
\caption{Distances in latent space between transformed samples. A random sample, here from the MNIST test set, is randomly transformed (here: rotated by small magnitues, in total between 20 and -20 degrees). The random sample and its transformed version are embedded by the learned encoder and the euclidean distance between these two embeddings is reported. In the graphs, $z_{\textrm{view}}$ refers to the embedding of the transformed version. $\alpha$ denotes the degree of rotation for generating the transformed sample. For each plot, we sample ten times from the variational posterior for every transformed embedding. (a) compares $\mathcal{T}_{\textrm{VAE}}$ and a standard VAE, while in (b) we compare $\mathcal{T}_{\textrm{VAE}}$ and VAE+.}
\label{fig:app:angle}
\end{center}
\vskip -0.2in
\end{figure}

\paragraph{Latent Space Structure.} The good results on the challenging out-of-sample KNN experiments lead us to first preliminary experiments that inspect the structure of the learned latent space in more detail: After training the model, we embed a sample in the latent space by using the encoder part of either $\mathcal{T}_{\textrm{VAE}}$, a standard VAE or VAE+. Then we transform this sample in observations space, embed this transformed version and compute the euclidean distance between this \emph{view} and the embedding of the original sample. In Figure~\ref{fig:app:angle} we see that the resulting distances are substantially smaller but also significantly less noisy for a $\mathcal{T}_{\textrm{VAE}}$. Note that for the reported experiment, the additive variant of a $\mathcal{T}_{\textrm{VAE}}$ is used, which is beneficial when considering euclidean distances in the latent space. We are currently running also experiments with the alternative $r_\chi$ versions. The transformation used in the Figure is a rotation (with an additional crop) between 20 and -20 degrees. 

\begin{figure}[t]
	\vskip 0.2in
    \begin{center}
	\includegraphics[width=.49\linewidth]{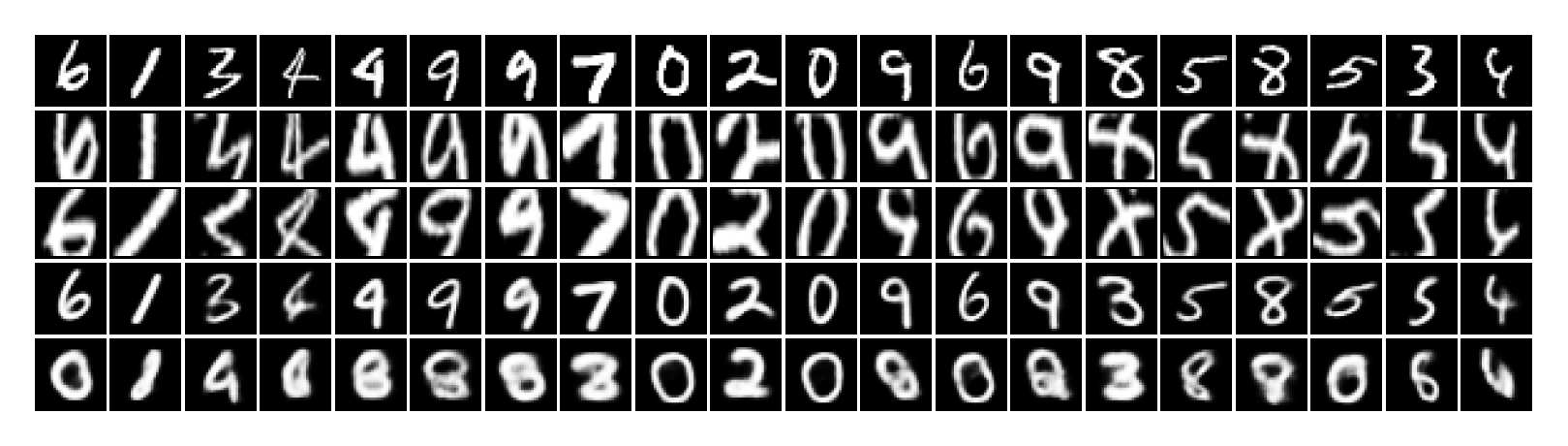}
	\includegraphics[width=.49\linewidth]{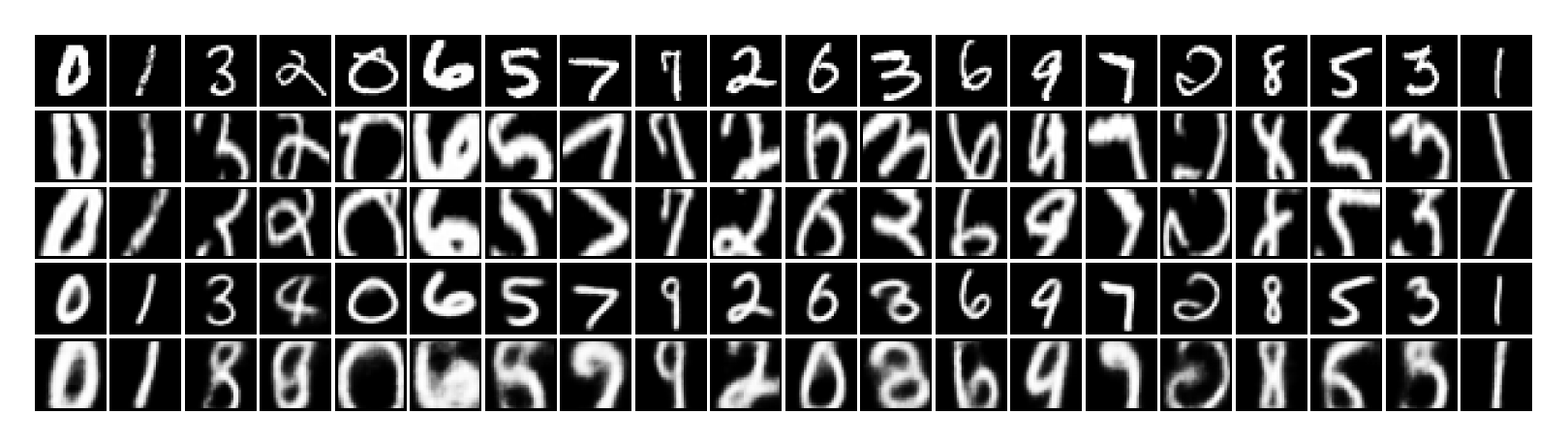}
	\includegraphics[width=.49\linewidth]{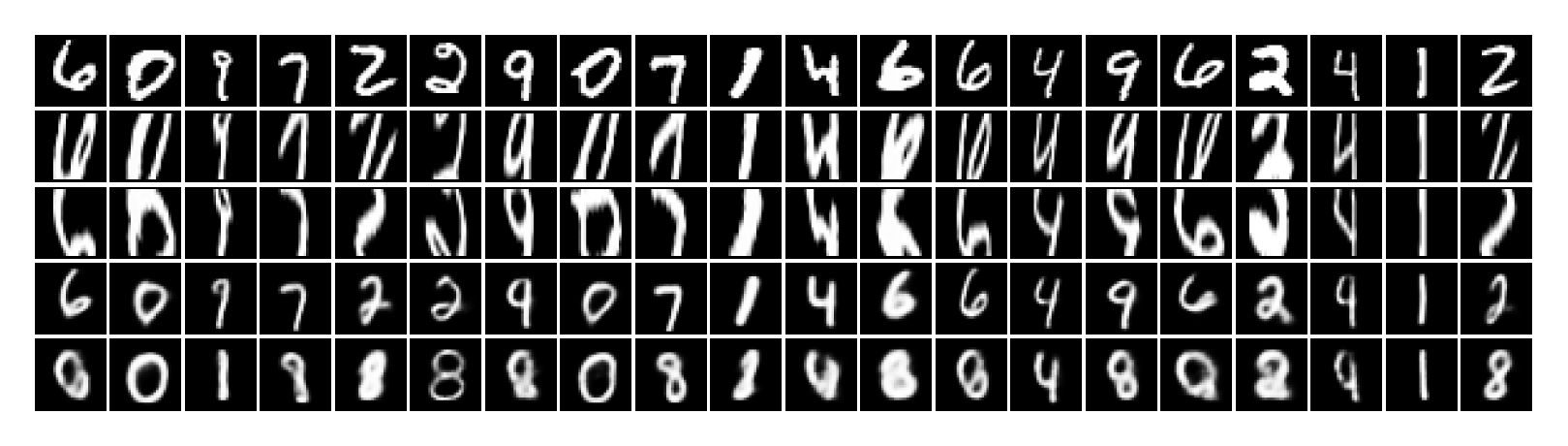}
	\includegraphics[width=.49\linewidth]{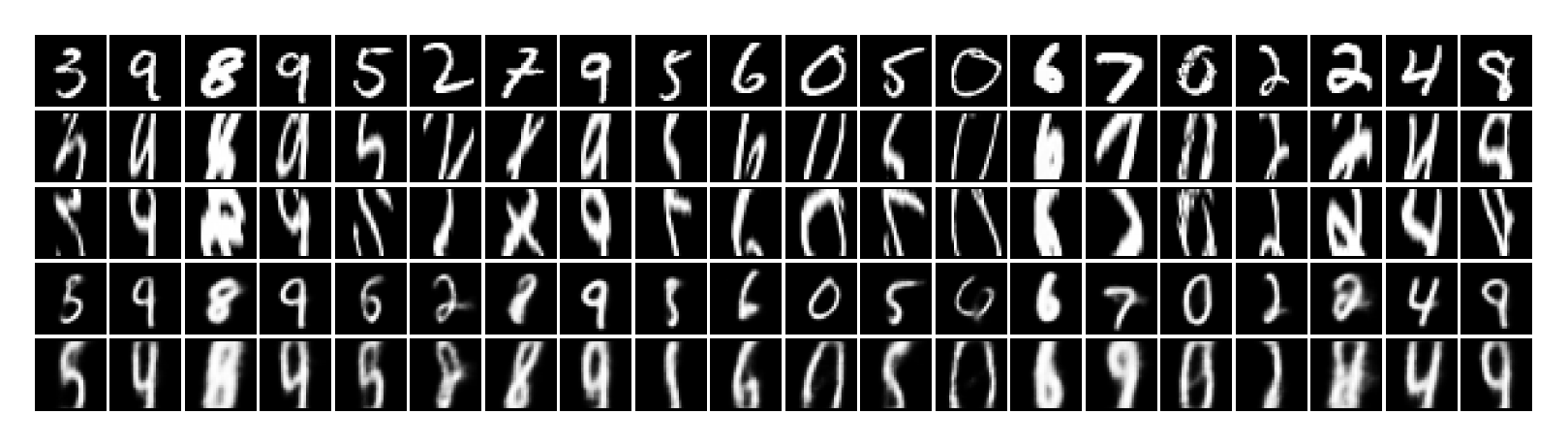}
	\includegraphics[width=.49\linewidth]{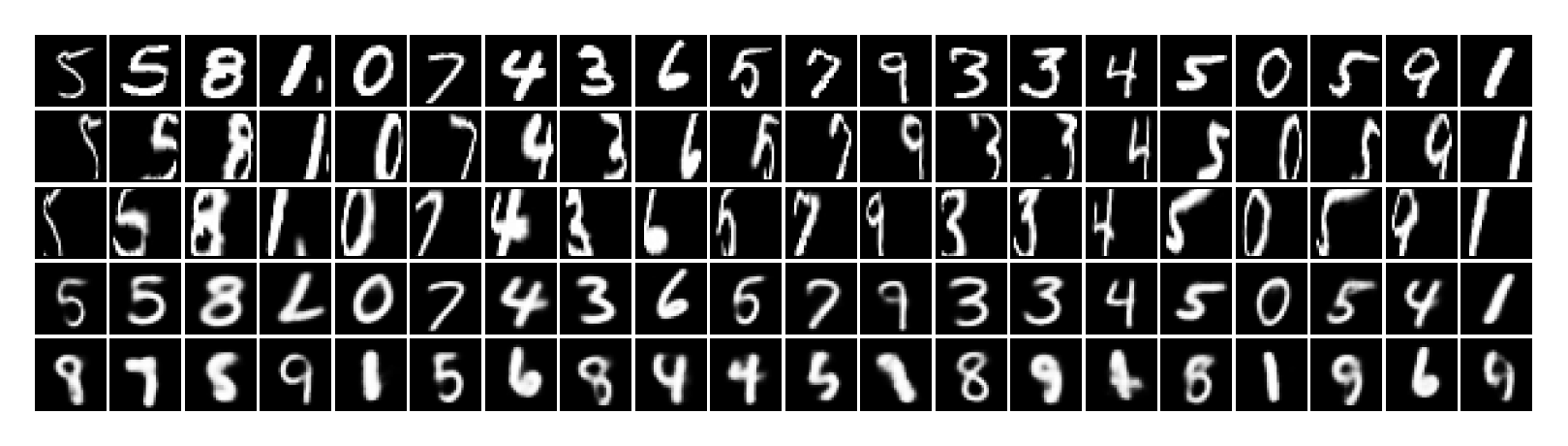}
	\includegraphics[width=.49\linewidth]{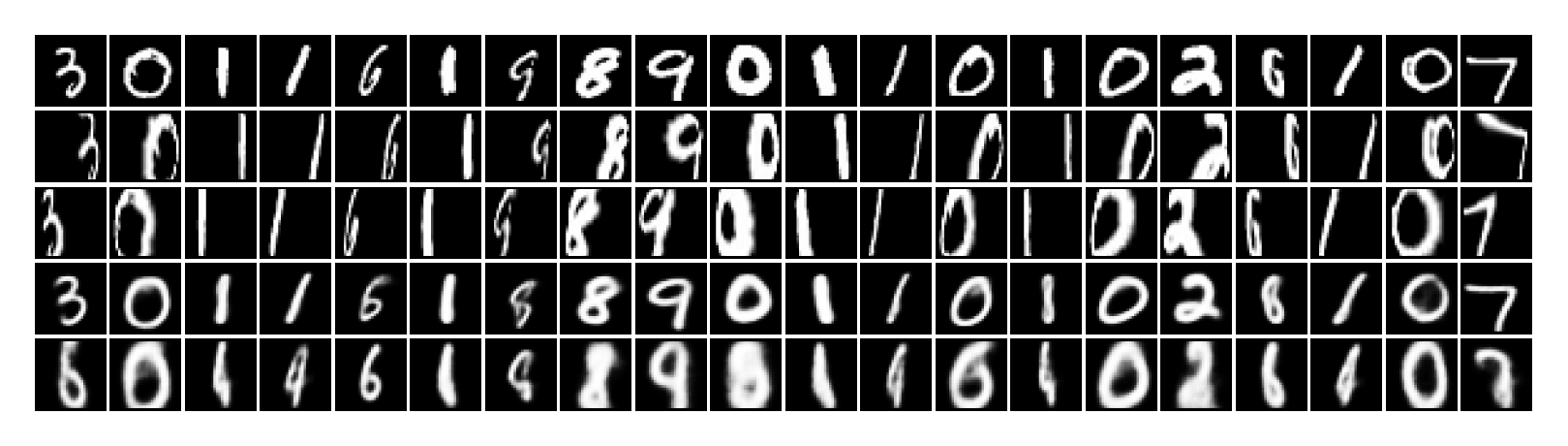}
	\includegraphics[width=.49\linewidth]{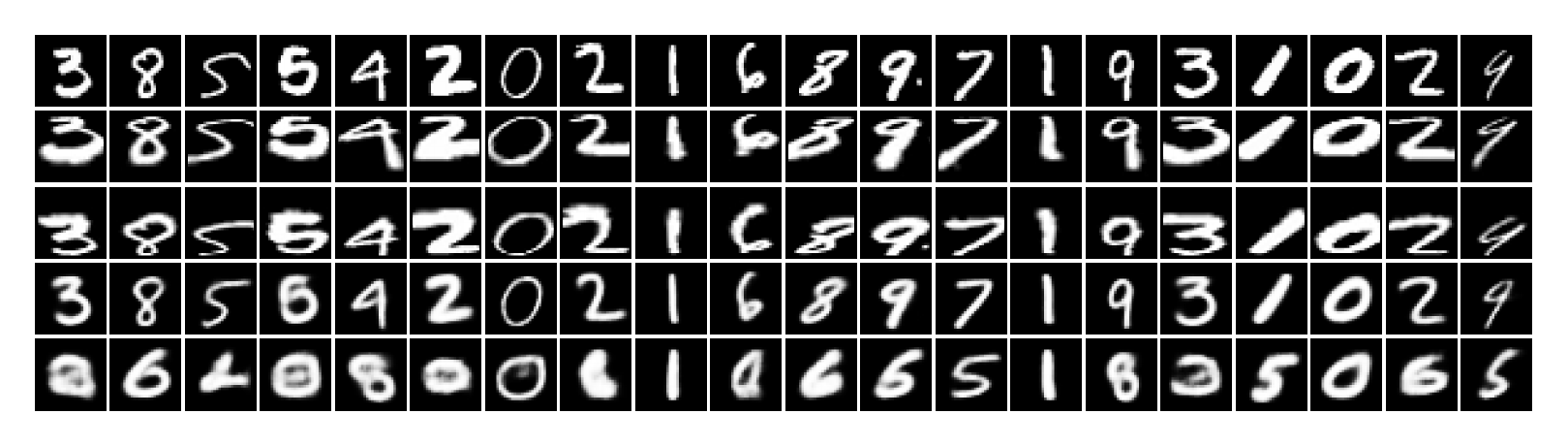}
	\includegraphics[width=.49\linewidth]{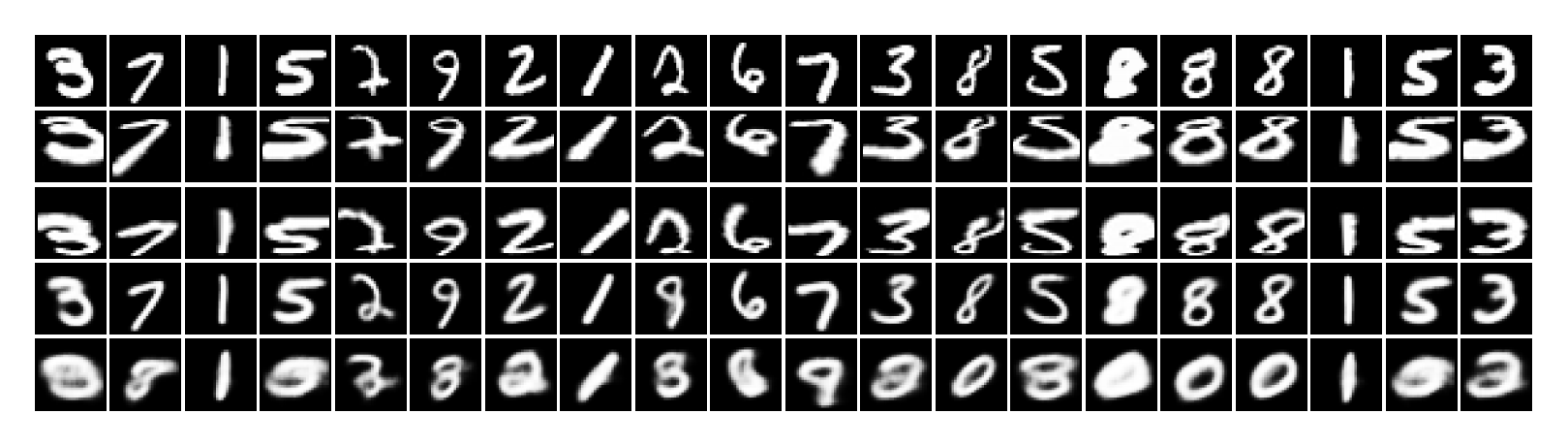}
	\caption{Transformations on MNIST. First row (per image): original sample. Second and third rows: views with applied transformations (resulting in $z_1$ and $z_2$). Fourth row: $\mathcal{T}_{\textrm{VAE}}$ interpolation reconstructions. Fifth row: VAE (set of images on the left side)/ VAE+ (right column) interpolation reconstructions. Interpolation is done by computing $z = \frac{z_1 + z_2}{2}$ and then reconstructing using the learned decoder. This setup is clearly helpful for the utilized function $r_{\chi}(z, \tau) \equiv z + \tau$ for $\mathcal{T}_{\textrm{VAE}}$.
	}
	\vskip - 0.2in
	\label{fig:app:mnist_figures}
	\end{center}
\end{figure}

\paragraph{Reconstruction properties -- MNIST.} In Figure~\ref{fig:app:mnist_figures} we explore the reconstruction properties of $\mathcal{T}_{\mathrm{VAE}}$ trained on MNIST \cite{lecun1998mnist} for different types of input transformations. We investigate qualitatively in what way the learned latent space has linear properties with respect to reconstruction results. This is best enforced by the additive version of $\mathcal{T}_{\textrm{VAE}}$ which we utilize here. The basic idea is to take a normal triplet (constructed according to the paper), embed the three samples, take the \emph{average} representation of the two embeddings representing the transformed samples and reconstruct using this average. Note that there is no apriori reason that this approach works well for either VAE or VAE+. They can not encode such a desired property in an easy way --- $\mathcal{T}_{\textrm{VAE}}$ can do so. This is nicely visible in the shown results and is true for all types of transformations.

\begin{figure}[t!]
\vskip 0.2in
\begin{center}
    \subfigure[$\mathcal{T}^{A}_{\mathrm{VAE}}$]{
    \includegraphics[width=.3\columnwidth]{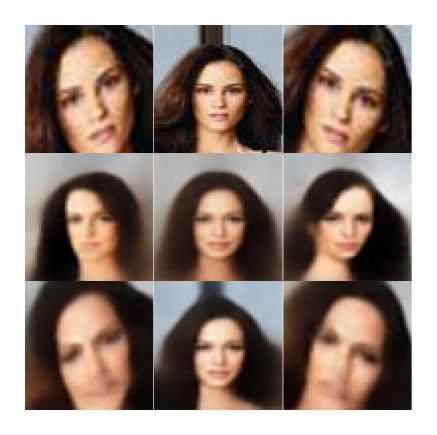}
    }
    \subfigure[VAE+]{
    \includegraphics[width=.3\columnwidth]{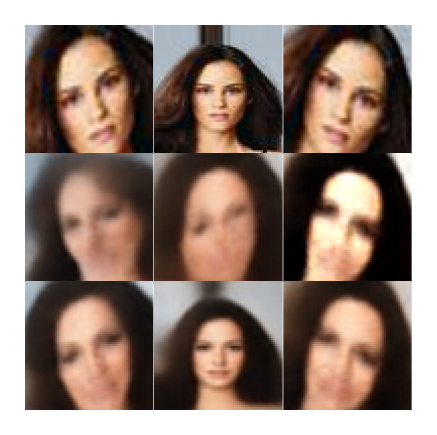}
    }
    \subfigure[VAE]{
    \includegraphics[width=.3\columnwidth]{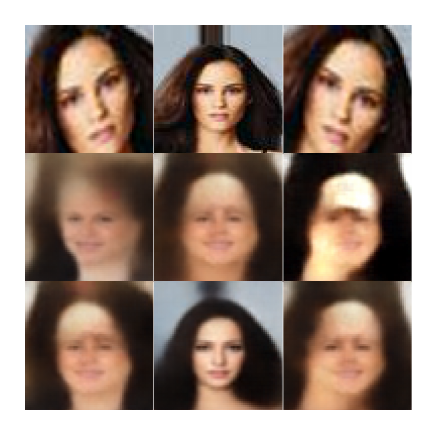}
    }
    \subfigure[$\mathcal{T}^{A}_{\mathrm{VAE}}$]{
    \includegraphics[width=.3\columnwidth]{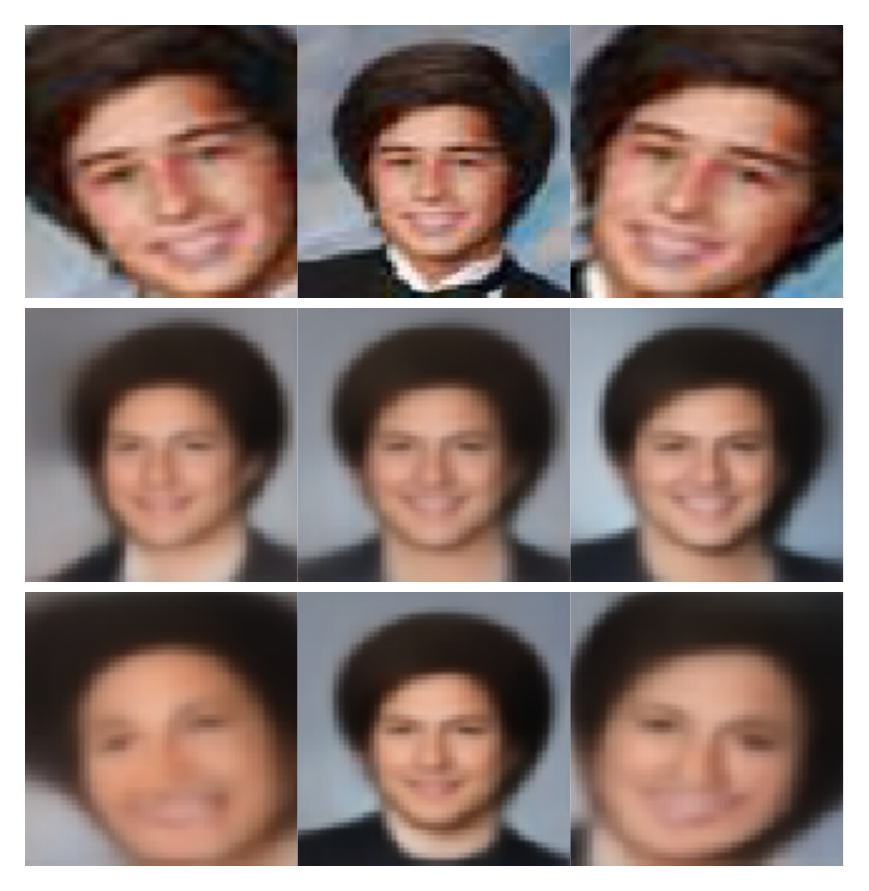}
    }
    \subfigure[VAE+]{
    \includegraphics[width=.3\columnwidth]{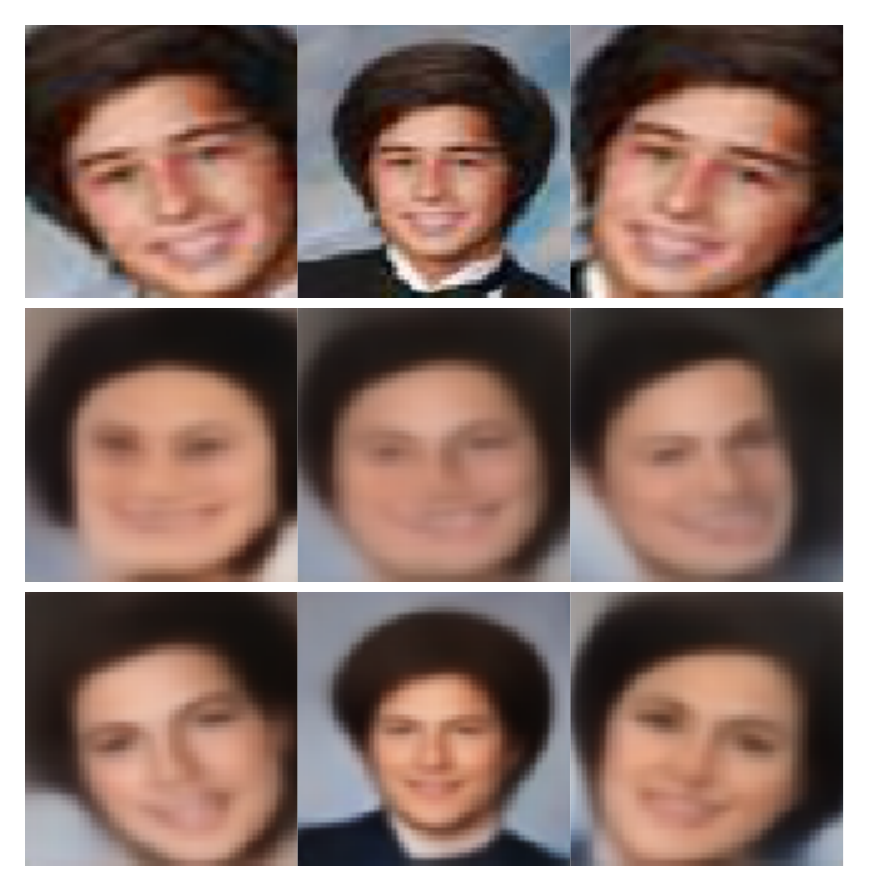}
    }
    \subfigure[VAE]{
    \includegraphics[width=.3\columnwidth]{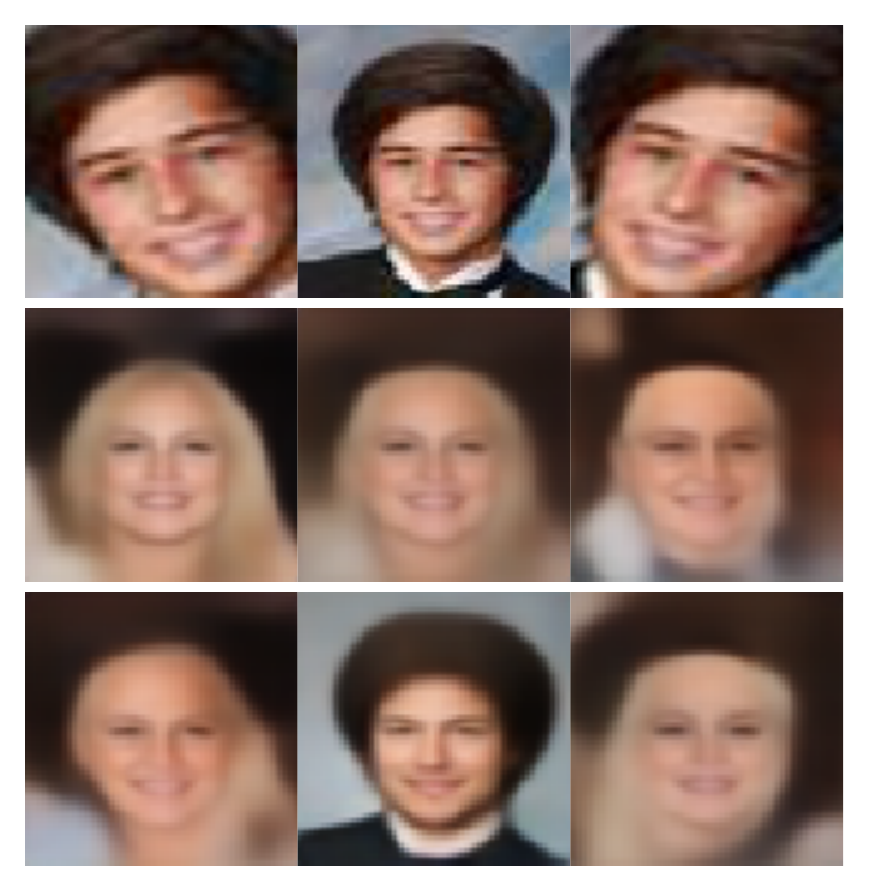}
    }
    \caption{Visualization for different models trained on CelebA. This qualitative experiment investigates how well linear interpolation works for the different models, $\mathcal{T}_{\mathrm{VAE}}$, VAE+ and VAE. The shown images are from the test set.}
\label{fig:operators}
\end{center}
\vskip -0.2in
\end{figure}

\paragraph{Reconstruction properties -- CelebA.} In Figure~\ref{fig:operators} we explore the reconstruction properties of $\mathcal{T}_{\mathrm{VAE}}$ when trained on the CelebA dataset \cite{liu2015faceattributes}. Every individual visualization in the Figure contains three rows: The first row is depicting the three samples forming a triplet, with the image in the middle being the original sample (that is, following our notation of the paper, the image order is the triplet $(x_1, x_0, x_2)$. After embedding the three images using a trained encoder, we can infer $z$ by $q_\psi(z_1, z_2$) for the $\mathcal{T}_{\mathrm{VAE}}$ or simply by $(z_1 + z_2)/2$ for a VAE and VAE+. The middle image in the second row is then reconstructed applying the learned decoder on this $z$. We can also infer the latent transformation -- either by $q_\xi(z_1, z_2)$ for the $\mathcal{T}_{\mathrm{VAE}}$ or simply with $\tau = |z_2 - z_1|/2$ for the two baseline models. The left and right image in the second row are then reconstructed similarly, either by applying the decoder to $r_\chi(z_1)$ or $r_\chi(z_2)$ respectively in the case of a $\mathcal{T}_{\mathrm{VAE}}$, or, for the two baselines, by applying the decoder to $z_0 + \tau$ and $z_0 - \tau$ respectively. The third row simply shows the reconstructions from the particular embeddings. The $\mathcal{T}_{\mathrm{VAE}}$ is trained with the additive variant, indicated by $\mathcal{T}_{\mathrm{VAE}}^{A}$ in the Figure. Again, the assumed linear structure in the latent space of a VAE and VAE+ is probably not correct at all, but there is no other principled way to get inter- or extrapolations for these two baseline models. The shown images are from the test set. Note that the transformed samples (left and right image in the first row), show a significantly different scale of the depicted image content --- this is due to the necessary cropping after the applied rotation transformation.

Clearly, a VAE struggles already reconstructing the transformations directly from the embeddings, having never seen these types of images. This is not a problem for VAE+. The VAE+ however can not reconstruct well from the interpolated latent embeddings. Specifically it can not handle the different scale of the content between the original sample and the transformed versions (see the middle image in the second row).  $\mathcal{T}_{\mathrm{VAE}}$ can handle this case very nice. It also can reconstruct the two transformed views (second row, left and right image) with respect to the content, but almost completely strips away the geometric component of these two views.

\paragraph{Reconstruction properties --- Failure Cases for $\mathcal{T}_{\mathrm{VAE}}$ on Omniglot.}
In the main paper we noticed how using the block-diagonal matrix variant (denoted by $M$) did not work properly for the experiment investigating the inferred latent transformations (cf. Figure 3 in the main text), even though all the metrics considered are compatible with the other variants, see for example Table 2 in the main text.

\begin{figure}[t!]
\vskip 0.2in
\begin{center}
    \subfigure[$\mathcal{T}^{A}_{\mathrm{VAE}}$]{
    \includegraphics[width=.3\columnwidth]{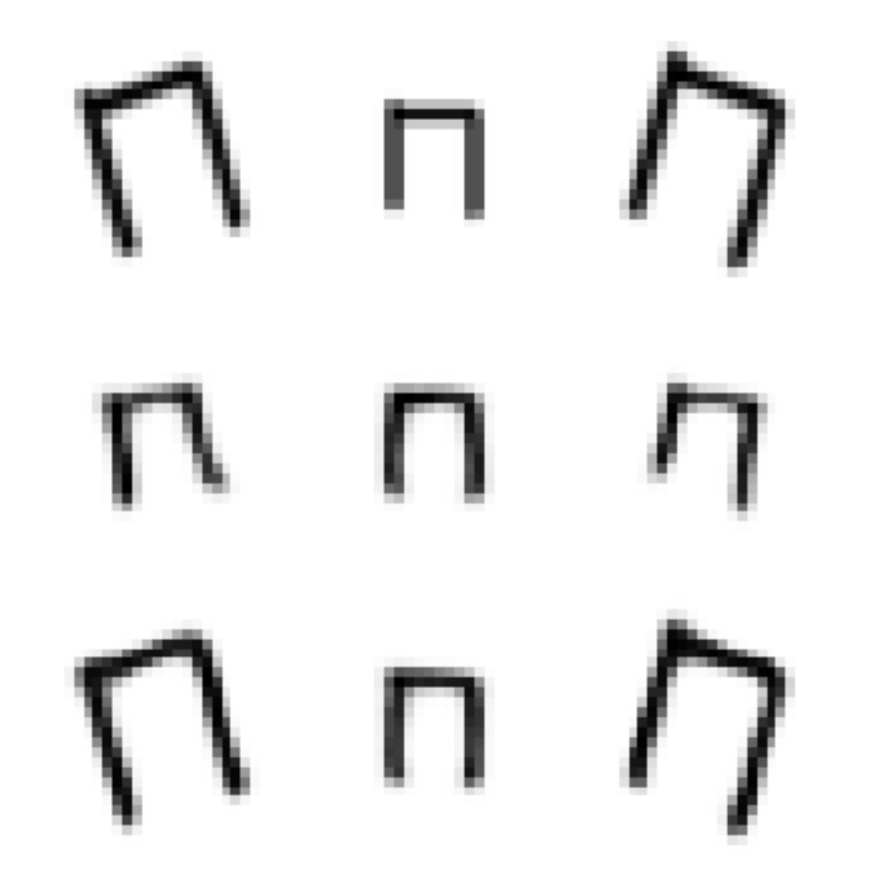}
    }
    \subfigure[$\mathcal{T}^{M}_{\mathrm{VAE}}$]{
    \includegraphics[width=.3\columnwidth]{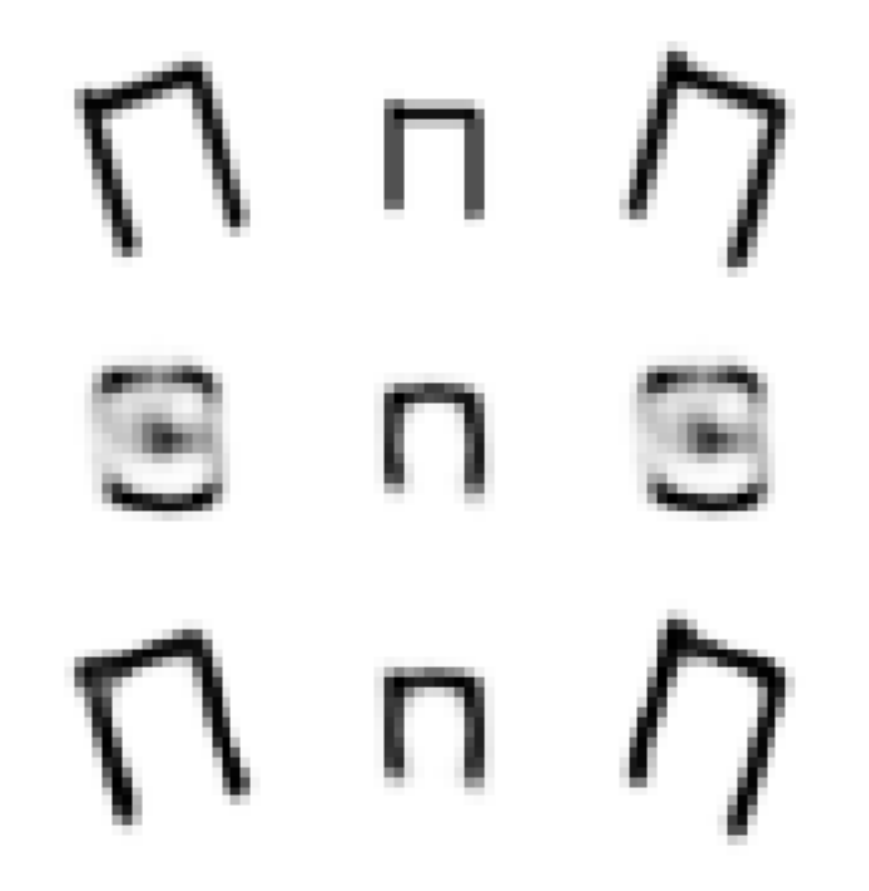}
    }
    \subfigure[$\mathcal{T}^{T}_{\mathrm{VAE}}$]{
    \includegraphics[width=.3\columnwidth]{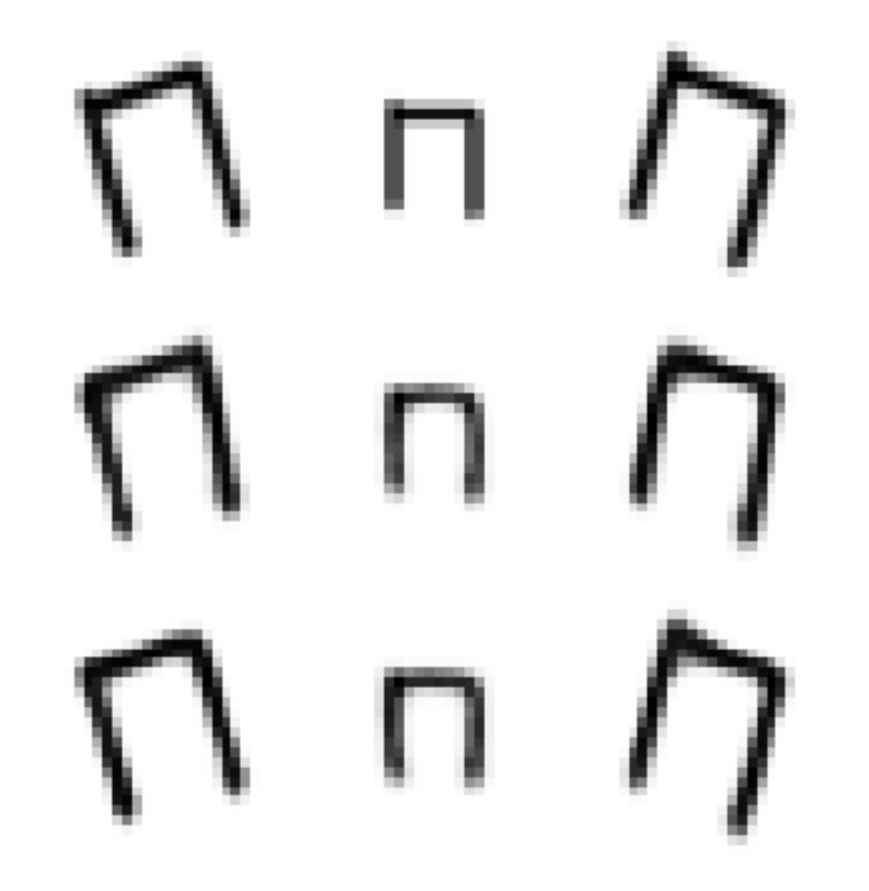}
    }
    \caption{Failure case for Omniglot. 
    Using the block diagonal formulation (Figure b) with 2d rotation matrices, we cannot represent transformations in latent space. We have a similar problem on all the considered datasets. Figure (a) shows the results for the additive version of $\mathcal{T}_\textrm{VAE}^A$ works reasonable well. Even better results can be obtained with a very free tridiagonal form, Figure (c).}
\label{fig:app:failure}
\end{center}
\vskip -0.2in
\end{figure}

In Figure~\ref{fig:app:failure}(b) we demonstrate this failure of the $M$ variant: The triplets in the first row are used to extrat their embeddings and the latent transformation $\tau$. It is then applied to the latent embedding $z_0$ in order to reconstruct the two transformed views, which does not work. Interestingly, a simple extension of the block-diagonal matrix to a full tri-diagonal matrix (Figure~\ref{fig:app:failure}(c) works very nice. It also shows a clear improvement over the additive schema from Figure~\ref{fig:app:failure}(a). Again, this is interesting because by just considering metrics (e.g. Table 1 later in this text), no difference between these two appears to exist.

\clearpage
\section{Tables and Figures}
\label{appendix:tables_figures}

In the following tables we report results for models trained on MNIST, Fashion-MNIST, and OMNIGLOT.
We report $\mathcal{T}_{\mathrm{VAE}}$ trained with different latent space dimensions $\text{ZDIM}$ and configurations of $r_{\chi}$:
\begin{itemize}
    \item Additive (A): $r_{\chi}(z_1 \vert z,\tau) = z +\tau$, $r_{\chi}(z_2 \vert z,\tau^{-1}) = z -\tau$.
    \item Matrix (M): $r_{\chi}(z_1 \vert z,\tau) = \tau \circ z$,  $r_{\chi}(z_2 \vert z,\tau^{-1}) =\tau^{-1} \circ z$, with $\tau$ a block-diagonal matrix, where each block is a 2d rotational matrix. $\circ$ represents matrix-vector multiplication.
    \item Matrix-Additive (MA): : $r_{\chi}(z_1 \vert z,\tau) = \tau \circ z + b$, $r_{\chi}(z_2 \vert z,\tau^{-1}) = \tau^{-1} \circ z + \tau^{-1} \circ b$, with $\tau$ a block-diagonal matrix, where each block is a 2d rotational matrix and $b$ a learned offset. 
    $\circ$ represents matrix-vector multiplication.
   \item Tridiagonal (T): $r_{\chi}(z_1 \vert z,\tau) = \tau \circ z + b$, $r_{\chi}(z_2 \vert z,\tau^{-1}) = \tau^{T} \circ z + \tau^{T} \circ b$ with $\tau$ a generic tridiagonal matrix and $b$ a learned offset. 
   $\circ$ denotes matrix-vector multiplication. We use $\tau^T$ to approximate the inverse matrix of $\tau$.
    \item Neural (N): $r_{\chi}(z_1 \vert z,\tau) = \text{NN}(z,\tau)$, $r_{\chi}(z_2 \vert z,\tau^{-1}) = \text{NN}(z, -\tau)$.
\end{itemize}

We also consider a residual variant $\mathcal{T}^{R}_{\mathrm{VAE}}$, where $r_{\chi} = z + r(z,\tau)$. That is, this variant is denoted by a superscript $R$ in the following tables.

For all the models we report:
\begin{itemize}
    \item[$-$] $\text{NLLX0}$, $\text{NLLX1}$, $\text{NLLX2}$ - reconstruction errors for $x_0$, $x_1$, $x_2$.
    \item[$-$] $\text{C1}$, $\text{C2}$ - MMDs on $z_1$ and $z_2$.
    \item[$-$] $\text{NLL}$ - Reconstructing $x_0$ from the inferred $z$. 
    \item[$-$] $\text{DIV}$ - Divergence on $z$.
    \item[$-$] $\text{KL0}$ - KL diverge on $z_0$.
    \item[$-$] $\text{ELBO}$ - Evidence Lower-Bound.
    \item[$-$] $\text{MLL}$ - Approximate log marginal likelihood.
    \item[$-$] $\text{KNN}_{in}$ - KNN accuracy in-sample test set.
    \item[$-$] $\text{KNN}_{out}$ - KNN accuracy out-of-sample AffNIST test set (100000 anchors, 320000 test samples).
\end{itemize}

$\mathcal{T}^{R}_{\mathrm{VAE}}$ performs comparably or better with respect to every considered metric.
In the main paper we report approximate log-marginal likelihood (MLL) on the original dataset for all the models, and VAE slightly outperforms $\mathcal{T}^{R}_{\mathrm{VAE}}$. From these tables, we can clearly see that $\mathcal{T}^{R}_{\mathrm{VAE}}$ performs a better reconstruction, and that the higher MLL is given by a slightly higher KL divergence (KL0). 
This result seems reasonable, because our model needs to account for a more challenging inference procedure.
For Omniglot we do not report the $\text{KNN}_{in}$ because the classes in the train set and test set are different and we should use a standard meta-learning evaluation instead of our KNN evaluation routine.

\begin{table}[h]
\caption{Metrics for MNIST}
\vskip 0.15in
\centering
\begin{small}
\begin{sc}
\begin{tabular}{lcccccccccccccc}
\toprule
model        & $r_{\chi}$       & zdim  & nllx0 & nllx1 & nllx2 & c1    & c2    & nll   & div   & kl0   & elbo  & mll   & $\text{knn}_{in}$ & $\text{knn}_{out}$ \\
\midrule
$\mathcal{T}_{\mathrm{VAE}}$     & A   & 100   & 70.59 & 93.52 & 92.14 & 9.13 & 9.14 & 72.31 & 25.35 & 25.94 & 96.53 & 91.61  & 0.98 & 0.85\\ 
$\mathcal{T}_{\mathrm{VAE}}$     & M  & 100    & 72.75 & 94.43 & 92.99 & 3.00 & 2.99 & 73.52 & 23.84 & 25.21 & 97.96 & 92.49 &0.98 & 0.83\\
$\mathcal{T}^{R}_{\mathrm{VAE}}$ & M  & 100    & 72.73 & 94.46 & 93.03 & 3.05 & 3.08 & 73.24 & 24.28 & 25.37 & 98.10 & 93.08 &0.98 & 0.85\\
$\mathcal{T}_{\mathrm{VAE}}$     & MA  & 100    & 71.81 & 92.95 & 91.60 & 1.54 & 1.62 & 72.82 & 23.41 & 24.83 & 96.65 & 91.84 &0.98 & 0.85\\
$\mathcal{T}^{R}_{\mathrm{VAE}}$ & MA  & 100    & 72.11 & 93.72 & 92.38 & 1.58 & 1.66 & 72.97 & 23.11 & 24.35 & 96.46 & 91.70 &0.98 & 0.84\\ 
$\mathcal{T}_{\mathrm{VAE}}$     & T   & 100   & 71.15 & 93.93 & 92.71 & 1.18 & 1.92 & 72.19 & 24.11 & 25.42 & 96.58 & 91.78  & 0.98 & 0.81\\ 
$\mathcal{T}^{R}_{\mathrm{VAE}}$ & T   &  100   & 71.33 & 93.88 & 92.55 & 1.04 & 1.05 & 73.04 & 23.28 & 27.66 & 98.99 & 91.92 &0.98 & 0.82\\ 
$\mathcal{T}_{\mathrm{VAE}}$     & N   & 100   & 71.45 & 95.68 & 94.22 & 0.78 & 0.62 & 72.94 & 23.36 & 24.58 & 96.03 & 91.72 & 0.98 & 0.81\\ 
$\mathcal{T}^{R}_{\mathrm{VAE}}$ & N   &  100   & 71.62 & 93.28 & 91.98 & 1.03 & 1.04 & 72.45 & 23.59 & 25.98 & 97.60 & 91.76 &0.98 & 0.81\\ 
VAE+    & -         & 100   & 71.95 & 112.13 & 110.55 & - & - & - & - & 24.13 & 96.09 & 92.16 & 0.98 & 0.75\\ 
VAE     & -         & 100   & 72.44 & - & - & - & - & - & - & 21.43 & 93.87 & 90.02 & 0.96 & 0.64\\ 
\midrule
$\mathcal{T}_{\mathrm{VAE}}$     & A  & 25    & 69.46 & 97.97 & 96.70 & 3.56 & 3.56 & 71.85 & 25.98 & 26.09 & 95.55 & 91.23  & 0.98 & 0.82\\ 
$\mathcal{T}_{\mathrm{VAE}}$     & M  & 26    & 71.69 & 97.49 & 96.00 & 3.32 & 3.31 & 72.82 & 24.18 & 24.42 & 96.11 & 90.98 &0.98 & 0.84\\
$\mathcal{T}^{R}_{\mathrm{VAE}}$ & M  & 26    & 71.66 & 97.48 & 96.24 & 3.49 & 3.51 & 72.46 & 24.48 & 24.34 & 96.01 & 91.14 &0.98 & 0.83\\
$\mathcal{T}_{\mathrm{VAE}}$     & MA  & 26    & 71.15 & 96.41 & 94.89 & 1.70 & 1.76 & 72.47 & 23.84 & 24.33 & 95.48 & 91.03 &0.98 & 0.83\\ 
$\mathcal{T}^{R}_{\mathrm{VAE}}$ & MA  & 26    & 70.97 & 97.19 & 95.78 & 1.66 & 1.77 & 72.56 & 23.71 & 24.49 & 95.46 & 90.60 &0.98 & 0.82\\
$\mathcal{T}_{\mathrm{VAE}}$     & T  & 25    & 70.15 & 97.42 & 96.06 & 1.26 & 1.32 & 71.96 & 24.69 & 25.00 & 95.15 & 90.54  & 0.98 & 0.81\\ 
$\mathcal{T}^{R}_{\mathrm{VAE}}$ & T  & 25    & 70.15 & 97.11 & 95.80 & 0.87 & 0.93 & 72.30 & 24.04 & 25.03 & 95.18 & 90.94 &0.98 & 0.82\\ 
$\mathcal{T}_{\mathrm{VAE}}$     & N  & 25    & 70.04 & 97.26 & 95.85 & 0.63 & 0.63 & 71.98 & 24.30 & 25.04 & 95.08 & 90.93  & 0.98 & 0.80\\ 
$\mathcal{T}^{R}_{\mathrm{VAE}}$ & N  & 25    & 69.97 & 96.79 & 95.17 & 0.58 & 0.55 & 72.31 & 24.04 & 24.85 & 94.83 & 90.52 &0.98 & 0.81\\ 
VAE+    & -          & 25    & 72.35 & 112.90 & 111.61 & - & - & - & - & 23.94 & 96.29 & 92.55 & 0.98 & 0.76\\ 
VAE     & -          & 25    & 74.67 & - & - & - & - & - & - & 20.28 & 94.95 & 91.04 & 0.96 & 0.58\\
\midrule
$\mathcal{T}_{\mathrm{VAE}}$     & A   & 10    & 76.99 & 116.06 & 114.54 & 1.38 & 1.42 & 79.66 & 20.52 & 21.01 & 98.00 & 94.11 & 0.97 & 0.62\\ 
$\mathcal{T}_{\mathrm{VAE}}$     & M   & 10    & 77.41 & 117.41 & 115.20 & 2.92 & 3.17 & 80.10 & 20.75 & 21.28 & 98.69 & 94.76 &0.97 & 0.63\\
$\mathcal{T}^{R}_{\mathrm{VAE}}$ & M   & 10    & 77.17 & 116.41 & 114.93 & 2.96 & 3.01 & 79.99 & 20.58 & 21.05 & 98.22 & 94.64 &0.97 & 0.64\\
$\mathcal{T}_{\mathrm{VAE}}$     & MA   & 10    & 77.38 & 116.93 & 115.19 & 1.08 & 1.08 & 79.85 & 20.13 & 20.41 & 97.79 & 93.94 &0.97 & 0.62\\
$\mathcal{T}^{R}_{\mathrm{VAE}}$ & MA   & 10    & 77.15 & 116.74 & 115.49 & 1.13 & 1.13 & 79.97 & 20.08 & 20.62 & 97.77 & 94.03 &0.97 & 0.61\\ 
$\mathcal{T}_{\mathrm{VAE}}$     & T   & 10    & 77.36 & 117.42 & 115.48 & 0.42 & 0.42 & 79.77 & 20.08 & 20.34 & 97.70 & 93.83 & 0.97 & 0.61\\ 
$\mathcal{T}^{R}_{\mathrm{VAE}}$ & T   & 10    & 76.88 & 116.25 & 114.03 & 0.44 & 0.44 & 79.42 & 20.15 & 20.28 & 97.16 & 93.48 &0.97 & 0.63\\ 
$\mathcal{T}_{\mathrm{VAE}}$     & N   & 10    & 76.87 & 116.26 & 114.23 & 0.29 & 0.26 & 79.50 & 19.88 & 20.23 & 97.10 & 93.28 & 0.97 & 0.59\\ 
$\mathcal{T}^{R}_{\mathrm{VAE}}$ & N   & 10    & 77.16 & 116.41 & 115.02 & 0.26 & 0.25 & 79.95 & 19.50 & 19.84 & 96.99 & 93.52 &0.97 & 0.59\\ 
VAE+    & -                            & 10    & 78.48 & 122.07 & 120.50 & - & - & - & - & 20.45 & 98.93 & 95.42 & 0.97 & 0.65\\ 
VAE     & -          & 10    & 76.03 & - & - & - & - & - & - & 19.61 & 95.64 & 91.83 & 0.96 & 0.57\\ 
\bottomrule
\end{tabular}
\end{sc}
\end{small}
\vskip -0.1in
\end{table}

\begin{table}[h]
\caption{Metrics for Fashion-MNIST}
\vskip 0.15in
\centering
\begin{small}
\begin{sc}
\begin{tabular}{lcccccccccccccc}
\toprule
model        & $r_{\chi}$       & zdim  & nllx0 & nllx1 & nllx2 & c1    & c2    & nll   & div   & kl0   & elbo  & mll   & $\text{knn}_{in}$ & $\text{knn}_{out}$ \\
\midrule
$\mathcal{T}_{\mathrm{VAE}}$     & A & 100   & 220.57 & 261.71 & 260.46 & 9.23 & 9.23 & 224.48 & 16.74 & 21.45 & 242.02 & 236.06 & 0.87 & 0.78\\ 
$\mathcal{T}_{\mathrm{VAE}}$     & M   & 100   & 222.14 & 262.98 & 261.36 & 2.45 & 2.43 & 225.09 & 15.27 & 21.44 & 243.58 & 237.52 &0.86 & 0.80\\  
$\mathcal{T}^{R}_{\mathrm{VAE}}$ & M & 100   & 222.26 & 261.74 & 259.73 & 2.39 & 2.35 & 225.03 & 15.20 & 21.39 & 243.64 & 237.33 &0.86 & 0.80\\
$\mathcal{T}_{\mathrm{VAE}}$     & MA   & 100   & 222.16 & 262.23 & 261.11 & 1.38 & 1.38 & 225.49 & 14.95 & 18.44 & 240.60 & 236.05 &0.87 & 0.77\\   
$\mathcal{T}^{R}_{\mathrm{VAE}}$ & MA & 100   & 222.26 & 262.37 & 260.91 & 1.41 & 1.44 & 225.53 & 14.78 & 18.71 & 240.97 & 238.10 &0.86 & 0.79\\
$\mathcal{T}_{\mathrm{VAE}}$     & T & 100   & 221.34 & 263.91 & 262.19 & 1.28 & 1.71 & 224.99 & 15.53 & 18.23 & 239.58 & 236.98 & 0.87 & 0.74\\ 
$\mathcal{T}^{R}_{\mathrm{VAE}}$ & T   & 100   & 221.32 & 263.61 & 261.98 & 1.12 & 1.13 & 225.31 & 15.29 & 19.67 & 240.98 & 238.53 &0.87 & 0.73\\ 
$\mathcal{T}_{\mathrm{VAE}}$     & N & 100   & 220.59 & 263.46 & 262.53 & 0.37 & 0.41 & 224.94 & 15.35 & 23.10 & 243.69 & 238.61 & 0.86 & 0.74\\ 
$\mathcal{T}^{R}_{\mathrm{VAE}}$ & N   & 100   & 221.37 & 263.73 & 262.28 & 0.94 & 0.94 & 224.95 & 15.25 & 20.78 & 242.15 & 237.91 &0.86 & 0.73\\ 
VAE+    & -          & 100   & 222.92 & 277.78 & 276.59 & - & - & - & - & 15.38 & 238.30 & 236.33 & 0.84 & 0.44\\ 
VAE     & -          & 100   & 222.85 & - & - & - & - & - & - & 13.43 & 236.28 & 234.24  & 0.84 & 0.33\\ 
\midrule
$\mathcal{T}_{\mathrm{VAE}}$ &  A & 25    & 219.53 & 264.76 & 263.29 & 3.78 & 3.78 & 223.90 & 17.96 & 18.66 & 238.18 & 235.38 & 0.86 & 0.73\\ 
$\mathcal{T}_{\mathrm{VAE}}$     &  M & 26   & 221.60 & 265.00 & 263.78 & 2.96 & 2.90 & 225.31 & 15.82 & 17.26 & 238.86 & 235.71 &0.86 & 0.74\\
$\mathcal{T}^{R}_{\mathrm{VAE}}$ &  M   & 26   & 221.60 & 264.33 & 262.91 & 2.95 & 2.91 & 225.01 & 15.92 & 17.43 & 239.03 & 235.81 &0.86 & 0.70\\   
$\mathcal{T}_{\mathrm{VAE}}$     &  MA & 26   & 220.70 & 264.70 & 263.21 & 1.30 & 1.32 & 224.67 & 15.55 & 18.04 & 238.74 & 235.06 &0.87 & 0.74\\ 
$\mathcal{T}^{R}_{\mathrm{VAE}}$ &  MA   & 26   & 221.08 & 264.58 & 263.06 & 1.31 & 1.33 & 224.88 & 15.40 & 16.65 & 237.73 & 234.79 &0.87 & 0.73\\ 
$\mathcal{T}_{\mathrm{VAE}}$     &  T   & 25    & 220.39 & 264.32 & 262.70 & 1.15 & 1.34 & 224.26 & 16.09 & 17.94 & 238.33 & 234.90 & 0.87 & 0.73\\ 
$\mathcal{T}^{R}_{\mathrm{VAE}}$ &  T   & 25    & 220.48 & 264.31 & 262.63 & 0.91 & 0.91 & 224.66 & 15.58 & 17.54 & 238.02 & 234.85 &0.86 & 0.71\\ 
$\mathcal{T}_{\mathrm{VAE}}$     &  N   & 25    & 220.44 & 265.76 & 264.33 & 0.53 & 0.51 & 224.93 & 15.47 & 17.17 & 237.61 & 235.13 & 0.86 & 0.68\\ 
$\mathcal{T}^{R}_{\mathrm{VAE}}$ &  N    & 25    & 220.31 & 265.31 & 264.15 & 0.50 & 0.48 & 224.75 & 15.61 & 17.62 & 237.92 & 234.80 &0.86 & 0.69\\ 
VAE+    & -           & 25    & 223.99 & 277.98 & 277.22 & - & - & - & - & 14.97 & 238.96 & 236.98 & 0.84 & 0.41\\ 
VAE     & -          & 25     & 222.73 & - & - & - & - & - & - & 13.39 & 236.12 & 234.14  & 0.83 & 0.34\\ 
\midrule
$\mathcal{T}_{\mathrm{VAE}}$ &  A & 10    & 221.25 & 271.50 & 269.88 & 1.78 & 1.74 & 225.84 & 15.86 & 16.52 & 237.77 & 235.47 & 0.84 & 0.48\\ 
$\mathcal{T}_{\mathrm{VAE}}$     & M   & 10    & 221.59 & 271.88 & 270.90 & 2.90 & 2.79 & 226.41 & 15.58 & 16.51 & 238.10 & 235.61 &0.84 & 0.48\\
$\mathcal{T}^{R}_{\mathrm{VAE}}$ & M   & 10    & 221.37 & 270.72 & 268.98 & 2.82 & 2.82 & 226.31 & 15.59 & 16.96 & 238.33 & 235.78 &0.84 & 0.48\\
$\mathcal{T}_{\mathrm{VAE}}$     & MA   & 10    & 221.35 & 271.34 & 269.98 & 1.05 & 1.08 & 226.25 & 15.39 & 16.25 & 237.59 & 235.37 &0.84 & 0.49\\
$\mathcal{T}^{R}_{\mathrm{VAE}}$ & MA   & 10     & 221.32 & 271.22 & 270.33 & 1.04 & 1.07 & 226.22 & 15.31 & 16.56 & 237.88 & 235.56 &0.85 & 0.47\\ 
$\mathcal{T}_{\mathrm{VAE}}$ &  T & 10    & 221.28 & 271.27 & 270.25 & 0.63 & 0.49 & 225.98 & 15.23 & 16.07 & 237.35 & 235.29 & 0.84 & 0.47\\
$\mathcal{T}^{R}_{\mathrm{VAE}}$ & T   & 10    & 221.35 & 271.32 & 270.35 & 0.61 & 0.60 & 225.98 & 15.18 & 16.09 & 237.44 & 235.15 &0.85 & 0.48\\
$\mathcal{T}_{\mathrm{VAE}}$ &  N & 10    & 221.13 & 271.04 & 269.84 & 0.33 & 0.37 & 225.85 & 14.86 & 16.14 & 237.27 & 235.08 & 0.84 & 0.46\\ 
$\mathcal{T}^{R}_{\mathrm{VAE}}$ & N   & 10    & 221.23 & 271.62 & 270.13 & 0.31 & 0.34 & 225.83 & 15.14 & 16.38 & 237.61 & 235.30 &0.84 & 0.49\\
VAE+    & -           & 10    & 224.04 & 278.58 & 277.44 & - & - & - & - & 14.81 & 238.85 & 236.89 & 0.83 & 0.41\\ 
VAE     & -           & 10    & 224.51 & - & - & - & - & - & - & 12.62 & 237.13 & 235.19   & 0.83 & 0.31\\
\bottomrule
\end{tabular}
\end{sc}
\end{small}
\vskip -0.1in
\end{table}

\begin{table}[h]
\caption{Metrics for Omniglot}
\vskip 0.15in
\centering
\begin{small}
\begin{sc}
\begin{tabular}{lcccccccccccccc}
\toprule
model        & $r_{\chi}$       & zdim  & nllx0 & nllx1 & nllx2 & c1    & c2  & nll   & div   & kl0   & elbo  & mll   & $\text{knn}_{in}$ & $\text{knn}_{out}$ \\
\midrule
$\mathcal{T}_{\mathrm{VAE}}$ &  A  & 100   & 96.15 & 107.30 & 110.09 & 9.74 & 9.78 & 108.81 & 27.69 & 41.14 & 137.30 & 131.62    & - & 0.69\\ 
$\mathcal{T}_{\mathrm{VAE}}$     & M    & 100   & 96.68 & 109.43 & 111.80 & 6.69 & 6.73 & 110.59 & 25.38 & 44.35 & 141.03 & 130.40 &- & 0.68\\
$\mathcal{T}^{R}_{\mathrm{VAE}}$ & M   & 100   & 97.36 & 108.15 & 110.38 & 6.45 & 6.48 & 110.59 & 25.25 & 47.49 & 144.85 & 137.24 &- & 0.66\\
$\mathcal{T}_{\mathrm{VAE}}$     & MA    & 100    & 97.59 & 108.30 & 110.93 & 3.94 & 4.37 & 110.98 & 25.09 & 43.05 & 140.65 & 133.52 &- & 0.71\\ 
$\mathcal{T}^{R}_{\mathrm{VAE}}$ & MA   & 100   & 98.00 & 109.74 & 112.31 & 4.38 & 4.70 & 110.16 & 26.00 & 37.42 & 135.43 & 130.01 &- & 0.70\\ 
$\mathcal{T}_{\mathrm{VAE}}$ &  T  & 100   & 94.55 & 107.40 & 109.72 & 2.57 & 4.68 & 109.58 & 25.93 & 57.79 & 152.34 & 136.69 & - & 0.69\\ 
$\mathcal{T}^{R}_{\mathrm{VAE}}$ & T    & 100   & 96.73 & 109.41 & 111.98 & 3.65 & 3.71 & 110.40 & 25.81 & 47.33 & 144.07 & 138.04 &- & 0.71\\ 
$\mathcal{T}_{\mathrm{VAE}}$     &  N  & 100   & 97.34 & 109.13 & 111.63 & 3.54 & 3.58 & 110.79 & 25.39 & 50.80 & 148.14 & 132.45    & - & 0.68\\ 
$\mathcal{T}^{R}_{\mathrm{VAE}}$ & N    & 100   & 96.32 & 107.09 & 109.67 & 3.72 & 3.66 & 109.19 & 26.70 & 55.75 & 152.08 & 139.64 &- & 0.69\\ 
VAE+   & -           & 100   & 107.48 & 138.12 & 141.80 & - & - & - & - & 23.09 & 130.57 & 125.90  & - & 0.47\\ 
VAE    & -           & 100   & 114.65 & - & - & - & - & - & - & 20.89 & 135.54 & 128.96    & - & 0.46\\ 
\midrule
$\mathcal{T}_{\mathrm{VAE}}$ &  A      & 25    & 97.13 & 117.18 & 120.05 & 3.69 & 3.72 & 109.01 & 27.31 & 37.42 & 134.55 & 127.96    & -  & 0.60\\ 
$\mathcal{T}_{\mathrm{VAE}}$ & M       & 26    & 98.64 & 117.41 & 120.13 & 4.77 & 4.88 & 110.32 & 25.73 & 35.42 & 134.06 & 127.36 &- & 0.57\\
$\mathcal{T}^{R}_{\mathrm{VAE}}$ & M   & 26    & 98.09 & 115.62 & 118.40 & 4.81 & 4.87 & 109.97 & 25.78 & 35.59 & 133.68 & 127.72 &- & 0.59\\
$\mathcal{T}_{\mathrm{VAE}}$ & MA      & 26    & 98.72 & 116.69 & 119.47 & 2.26 & 2.42 & 109.57 & 25.65 & 34.34 & 133.06 & 127.13 &- & 0.62\\
$\mathcal{T}^{R}_{\mathrm{VAE}}$ & MA  & 26    & 97.91 & 116.27 & 118.90 & 2.25 & 2.50 & 109.54 & 25.64 & 36.25 & 134.16 & 127.45 &- & 0.59\\ 
$\mathcal{T}_{\mathrm{VAE}}$ &  T  & 25    & 99.00 & 117.71 & 120.38 & 1.46 & 1.58 & 109.95 & 26.18 & 33.30 & 132.30 & 127.51    & - & 0.59\\ 
$\mathcal{T}^{R}_{\mathrm{VAE}}$ & T    & 25    & 97.70 & 117.27 & 120.31 & 1.63 & 1.65 & 109.66 & 25.92 & 36.39 & 134.09 & 128.27 &- & 0.63\\ 
$\mathcal{T}_{\mathrm{VAE}}$ &  N  & 25    & 98.49 & 118.10 & 120.97 & 1.34 & 1.30 & 110.44 & 25.64 & 35.26 & 133.76 & 128.25    & - & 0.58\\ 
$\mathcal{T}^{R}_{\mathrm{VAE}}$ & N    & 25    & 99.06 & 117.54 & 120.34 & 1.20 & 1.16 & 109.64 & 25.96 & 34.59 & 133.65 & 127.32 &- & 0.59\\ 
VAE+    & -          & 25    & 108.90 & 139.61 & 143.25 & - & - & - & - & 22.78 & 131.68 & 126.60  & - & 0.50\\ 
VAE     & -          & 25    & 116.44 & - & - & - & - & - & - & 20.64 & 137.08 & 130.12    & - & 0.37\\ 
\midrule
$\mathcal{T}_{\mathrm{VAE}}$     & A   & 10    & 115.72 & 150.13 & 154.86 & 1.48 & 1.52 & 123.23 & 18.38 & 22.75 & 138.47 & 134.03   & - & 0.34\\ 
$\mathcal{T}_{\mathrm{VAE}}$     & M   & 10    & 115.81 & 150.45 & 154.77 & 2.95 & 2.97 & 123.62 & 18.59 & 22.62 & 138.44 & 133.70 &- & 0.38\\ 
$\mathcal{T}^{R}_{\mathrm{VAE}}$ & M   & 10    & 115.48 & 149.29 & 154.02 & 2.94 & 2.99 & 123.45 & 18.45 & 23.28 & 138.76 & 133.85 &- & 0.35\\ 
$\mathcal{T}_{\mathrm{VAE}}$     & MA   & 10    & 115.42 & 149.63 & 154.00 & 1.09 & 1.13 & 123.49 & 18.07 & 22.81 & 138.23 & 133.30 &- & 0.34\\  
$\mathcal{T}^{R}_{\mathrm{VAE}}$ & MA   & 10    & 116.51 & 150.68 & 155.02 & 0.97 & 0.98 & 124.55 & 18.06 & 22.64 & 139.15 & 134.71 &- & 0.38\\ 
$\mathcal{T}_{\mathrm{VAE}}$     & T   & 10    & 115.59 & 150.37 & 154.84 & 0.51 & 0.51 & 122.99 & 18.30 & 22.02 & 137.62 & 133.75   & - & 0.33\\ 
$\mathcal{T}^{R}_{\mathrm{VAE}}$ & T   & 10    & 116.05 & 150.82 & 155.54 & 0.65 & 0.71 & 124.11 & 18.07 & 22.75 & 138.80 & 133.79 &- & 0.40\\ 
$\mathcal{T}_{\mathrm{VAE}}$     & N   & 10    & 115.83 & 150.35 & 154.89 & 0.47 & 0.49 & 123.51 & 17.88 & 22.59 & 138.43 & 133.86   & - & 0.42\\ 
$\mathcal{T}^{R}_{\mathrm{VAE}}$ & N   & 10    & 116.07 & 149.61 & 154.41 & 0.44 & 0.44 & 123.24 & 18.27 & 22.26 & 138.33 & 134.09 &- & 0.36\\ 
VAE+    & -          & 10    & 117.90 & 154.20 & 159.00 & - & - & - & - & 17.95 & 135.85 & 131.74  & - & 0.33\\ 
VAE     & -          & 10    & 122.32 & - & - & - & - & - & - & 18.10 & 140.43 & 133.77    & - & 0.32\\  
\bottomrule
\end{tabular}
\end{sc}
\end{small}
\vskip -0.1in
\end{table}
\clearpage
\section{Technical Details}
\label{appendix:details}

We used the same setting to train on MNIST \citep{lecun1998mnist}, Fashion-MNIST \citep{xiao2017fashion}, and Omniglot \citep{lake2015human}.  All the models are trained with the same architecture, hyper-parameters andtransformations.
All the datasets were resized to $28 \,\text{x}\, 28$.

We used mini-batches of size 100 and trained the models for 200 epochs, using the Adam optimizer with a learning rate of $\alpha=10^{-4}$. \citep{kingma2014adam}.  Every 50 epochs, we halved $\alpha$. We used $4\,\text{x}\,4$ filters for all the encoder and decoder layers and strides of 2 for all the encoder convolutions and the last two transposed in the decoder\footnote{We write for all the architectures  $\textrm{Module}_{k}$ where $k$ is the number of filters in output for convolutions and transpose convolutions and number of units in output for fully connected layers.}.

The encoder parameterizes the moments of a multivariate Gaussian distribution with a diagonal covariance matrix.
The decoder parameterizes the moments of a Bernoulli distribution.
We trained the models with a binary cross-entropy loss and minimized the KL divergence with standard normal as prior.
We trained models with latent dimensions $z_{d} \in \left[10, 25, 100\right]$.

$q_{\phi}(z \vert x)$:
\begin{align*}
    x \in \mathcal{R}^{28 \text{x} 28} &\rightarrow \text{Conv}_{32} \rightarrow \text{ReLU}         \\  
                                       &\rightarrow \text{Conv}_{32}        \rightarrow \text{ReLU}  \\  
                                       &\rightarrow \text{Conv}_{64}        \rightarrow \text{ReLU}  \\  
                                       &\rightarrow \text{Conv}_{128}       \rightarrow \text{ReLU}  \\  
                                       &\rightarrow \text{Conv}_{2 z_{d}}                                
\end{align*}

$p_{\theta}(x \vert z)$:
\begin{align*}
    z \in \mathcal{R}^{z_{d}}   &\rightarrow \text{Conv}_{128} \rightarrow \text{ReLU} \\
                                &\rightarrow \text{ConvT}_{64}  \rightarrow \text{ReLU} \\
                                &\rightarrow \text{ConvT}_{32}  \rightarrow \text{ReLU} \\
                                &\rightarrow \text{ConvT}_{32}  \rightarrow \text{ReLU} \\
                                &\rightarrow \text{ConvT}_{1}
\end{align*}

$q_{\psi}(z \vert z_1, z_2)$:
\begin{align*}
    z \in \mathcal{R}^{2 z_{d}}   &\rightarrow \text{FC}_{1000} \rightarrow \text{ReLU} \\
                                &\rightarrow \text{FC}_{1000} \rightarrow \text{ReLU} \\
                                &\rightarrow \text{FC}_{z_{d}}
\end{align*}

$\tau_{\xi}(z_1, z_2)$:
\begin{align*}
    z \in \mathcal{R}^{2 z_{d}}   &\rightarrow \text{FC}_{1000} \rightarrow \text{ReLU} \\
                                &\rightarrow \text{FC}_{1000} \rightarrow \text{ReLU} \\
                                &\rightarrow \text{FC}_{z_{d}}.
\end{align*}

The neural version of $r_{\chi}$ has the same architecture as $\tau_\xi(z_1, z_2)$, it also takes.
For CelebA \citep{liu2015faceattributes} we use a reduced version of the VAE architecture and training procedure proposed in \citep{dai2019diagnosing}, with a gaussian observation model and fixed variance.

\end{document}